\documentclass{article}
\pdfoutput=1

\usepackage[final, nonatbib]{neurips_2024}

\usepackage{algorithmic}
\usepackage{textcomp}
\usepackage{color}
\usepackage{colortbl}
\usepackage[table,xcdraw]{xcolor}        
\usepackage{wrapfig}
\definecolor{citecolor}{RGB}{34,139,34}
\usepackage[colorlinks,linkcolor=red,citecolor=citecolor,urlcolor=blue]{hyperref}

\usepackage{amsthm}
\usepackage{pifont}     

\newtheorem{theorem}{Theorem}[section]

\newtheorem{corollary}[theorem]{Corollary}
\usepackage{nicefrac}       
\usepackage[utf8]{inputenc} 
\usepackage[T1]{fontenc}    
\usepackage{booktabs}       
\usepackage{graphicx}

\usepackage{threeparttable}

\usepackage{multirow}

\usepackage{amsmath}
\usepackage{bm}
\usepackage{bbm}
\usepackage{amsthm}
\usepackage{amsfonts}

\usepackage{enumitem} 
\usepackage[subrefformat=parens,labelformat=parens]{subfig}

\usepackage{wrapfig}
\usepackage[ruled, vlined, linesnumbered, noend]{algorithm2e}

\renewcommand{\vec}[1]{\boldsymbol{#1}}

\theoremstyle{plain}

\theoremstyle{definition}

\newcommand{\eps}{\epsilon}

\newcommand{\neurolight}{\texttt{NeurOLight}\xspace}
\newcommand{\ffno}{\texttt{NeuroLight}\xspace}

\newcommand{\name}{\texttt{PACE}\xspace}
\newcommand{\namea}{\texttt{PACE-I}\xspace}
\newcommand{\nameb}{\texttt{PACE-II}\xspace}

\newcommand{\VE}{\mathbf{E}}
\newcommand{\Ex}{\mathbf{E}_x}
\newcommand{\Ey}{\mathbf{E}_y}
\newcommand{\Ez}{\mathbf{E}_z}
\newcommand{\J}{\mathbf{J}}
\newcommand{\VH}{\mathbf{H}}
\newcommand{\Hx}{\mathbf{H}_x}
\newcommand{\Hy}{\mathbf{H}_y}
\newcommand{\Hz}{\mathbf{H}_z}
\newcommand{\curl}{\nabla\times}

\newcommand{\normx}{\hat{\mathbf{x}}}
\newcommand{\normy}{\hat{\mathbf{y}}}
\newcommand{\normz}{\hat{\mathbf{z}}}

\renewcommand{\vec}[1]{\boldsymbol{#1}} 

\usepackage{pifont}

\title{PACE: \underline{P}acing Operator Learning to \underline{Ac}curate Optical Field Simulation for \underline{C}omplicated Photonic D\underline{e}vices}

\author{
Hanqing Zhu\textsuperscript{\ding{170} $\dagger$}, Wenyan Cong\textsuperscript{\ding{170}}, Guojin Chen\textsuperscript{\ding{170}}\thanks{This work was done when Guojin Chen was a visiting scholar at UT Austin.}, Shupeng Ning\textsuperscript{\ding{170}}, \\ 
\textbf{Ray T. Chen\textsuperscript{\ding{170}}, Jiaqi Gu\textsuperscript{\ding{169}}, David Z. Pan\textsuperscript{\ding{170} $\ddagger$}} \\
\textsuperscript{\ding{169}}Arizona State University\\
\textsuperscript{\ding{170}}The University of Texas at Austin 
\\
\texttt{\textsuperscript{$\dagger$}hqzhu@utexas.edu, \textsuperscript{$\ddagger$}dpan@ece.utexas.edu}
}

\begin{document}
\maketitle

\begin{abstract}

Electromagnetic field simulation is central to designing, optimizing, and validating photonic devices and circuits. 
However, costly computation associated with numerical simulation poses a significant bottleneck, hindering scalability and turnaround time in the photonic circuit design process.
Neural operators offer a promising alternative, but existing SOTA approaches,\neurolight, struggle with predicting high-fidelity fields for real-world \emph{complicated} photonic devices, with the best reported 0.38 normalized mean absolute error in \neurolight.
The interplays of highly complex light-matter interaction, e.g., scattering and resonance, sensitivity to local structure details, non-uniform learning complexity for full-domain simulation, and rich frequency information, contribute to the failure of existing neural PDE solvers.
In this work, we boost the prediction fidelity to an unprecedented level for simulating complex photonic devices with a novel operator design driven by the above challenges.
We propose a novel cross-axis factorized \name operator with a strong long-distance modeling capacity to connect the full-domain complex field pattern with local device structures.
Inspired by human learning, we further divide and conquer the simulation task for extremely hard cases into two progressively easy tasks, with a first-stage model learning an initial solution refined by a second model.
On various \emph{complicated} photonic device benchmarks, we demonstrate one sole \name model is capable of achieving \textbf{73\%} lower error with \textbf{50\%} fewer parameters compared with various recent ML for PDE solvers.
The two-stage setup further advances high-fidelity simulation for even more intricate cases.
In terms of runtime, 
\name demonstrates \textbf{154-577}$\times$ and \textbf{11.8-12}$\times$ simulation speedup over numerical solver using scipy or highly-optimized pardiso solver, respectively.
\textbf{We open sourced the code and \emph{complicated} optical device dataset at \href{https://github.com/zhuhanqing/PACE-Light}{PACE-Light}.}

\end{abstract}
\section{Introduction}
\label{sec:Introduction}

With advances in integrated photonics,
photonic structures capable of transmitting or processing information are gathering increasing interest, fueled by the optical communication~\cite{shi2022silicon} and the recent resurgence of photonic analog computing~\cite{NP_Nature2021_Feldmann, NP_JLT24_Ning,NP_NaturePhotonics2021_Shastri, NP_Nature2021_Xu, NP_HPCA24_Zhu, zhu2022fuse, NP_ACSPhotonics2022_Feng}.
Light-empowered communication and computing offer a promising pathway for reshaping future AI systems, prompting the optical community to discover compact, customized devices~\cite{gu2022squeezelight, wang2022integrated, NP_NatureComm2022_Zhu, feng2024integrated} to overcome the limitations of bulky optical components.
In this optical design process,
numerical simulators, e.g., the popular finite difference frequency domain (FDFD) algorithm~\cite{NP_ACSPhotonics2018_Lim}, is heavily used to obtain accurate optical fields for characterizing and optimizing device behavior.
However, the significant time and computational costs associated with Maxwell partial differential equation (PDE) simulations, exacerbated by the need for finely tailored meshes and numerous simulation runs for iterative optimization, 
pose substantial bottlenecks in the design loop.

\begin{figure}
    \centering
    \includegraphics[width=0.95\linewidth]{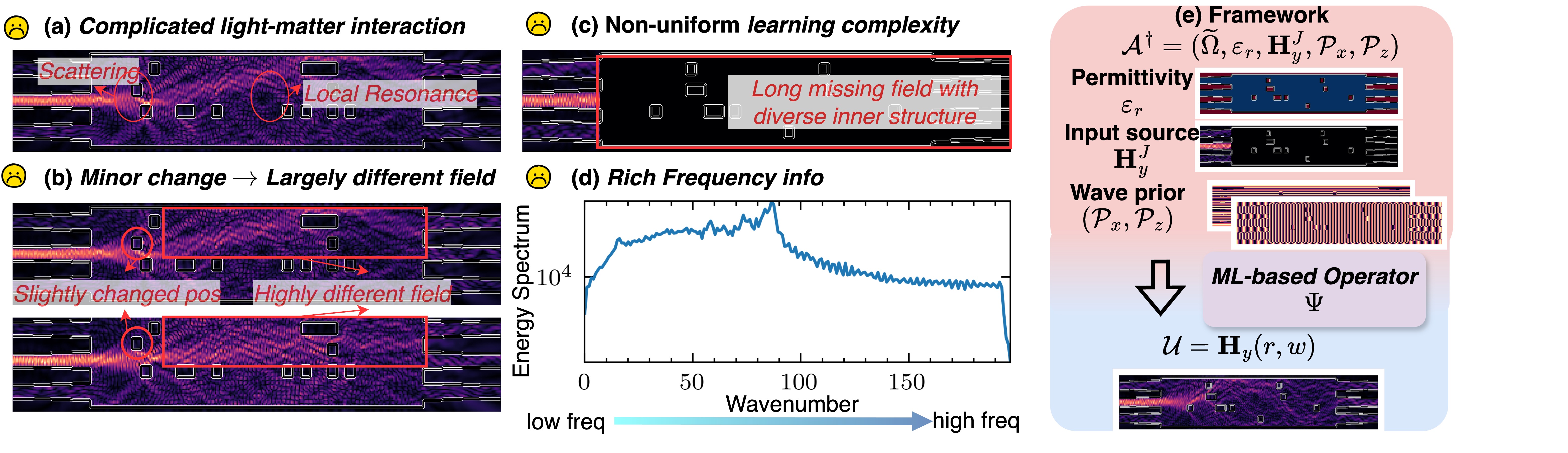}
    \caption{\small Challenges of complicated optical device simulation: (a-d) and learning framework (e).}
    \label{fig:Teaser}
    \vspace{-20pt}
\end{figure}

Recently, neural PDE solvers~\cite{cao2021choose, NP_Neurips2022_Gu,  NN_NeurIPS2021_Gupta, ML_Neurips2024_Li, NN_ICLR2021_Li, ML_ICLR2023_Tran,  ML_ICML2023_Wu} have emerged as promising surrogate models for \emph{fast} and \emph{accurate} PDE solving. 
\neurolight~\cite{NP_Neurips2022_Gu} represents the state-of-the-art (SOTA), extending neural operators to parametric photonic device simulations in a physics-agnostic manner.
However, it still exhibits large errors in simulating real-world \textit{complicated} optical devices, reporting a 0.38 normalized mean absolute error on the etched multi-mode interference (MMI) device~\cite{NP_SciRep2019_Tahersima}.
One may wonder what the major challenges are, given the successes of neural operators in many scientific PDEs.
Firstly, for \emph{complicated} devices, the permittivity distribution is discrete and highly contrasting, transforming the Maxwell PDE into a multi-scale problem~\cite{ammari2012multi}, further leading to complex light-matter interactions such as scattering and resonance, as illustrated in Fig.~\ref{fig:Teaser}(a).
Secondly, their optical fields are highly sensitive to local structural changes; even minor alterations can significantly impact the field, as depicted in Fig.~\ref{fig:Teaser}(b).
Moreover, with diversifying field patterns along the light propagation path, it shows non-uniform learning complexity especially in regions distant from the input light source.
Finally, a spectral analysis provides insights into the frequency-domain challenges, as illustrated in Fig.~\ref{fig:Teaser}(d). Unlike simpler systems where low frequencies dominate (e.g., Darcy flow shown in Fig.~\ref{fig:darcy}), \textit{complicated} devices exhibit rich frequency spectra with high-frequency components. This diversity underpins the difficulty faced by previous neural PDE solvers in accurately simulating \textit{complicated} photonic devices, supporting the assertion in~\cite{ML_Neurips2024_Li} that no single model can universally solve all types of PDEs.

In this work, we tackle the challenging\textit{ real-world complicated} optical device simulation problem.
We vastly boost prediction fidelity and keep \textbf{154-577}$\times$ and \textbf{11.8-12}$\times$ speedup over traditional numerical solver~\cite{NP_ACSPhotonics2018_Lim} on a 20-core CPU with \texttt{scipy} or highly-optimized \texttt{pardiso} solver, respectively. 

Overall, we make the following key contributions:
\begin{itemize}
    \item We introduce a novel cross-axis factorized \name operator backbone, effectively capturing complex physical phenomena across the full domain in a parameter-efficient manner.
    \item We employ a divide-and-conquer approach inspired by human learning for extremely challenging cases, with a first-stage \namea to learn a rough approximation of the optical field, refined by a second-stage \nameb.
    \item On various \textit{complicated} device benchmarks, one sole \name significantly outperforms baselines, achieving \textbf{73\%} lower error with \textbf{50\%} fewer parameters. Even compared to the best baseline, it lowers prediction error by over \textbf{39\%} with \textbf{17\%} fewer parameters.
    Our two-stage method further advances high-fidelity simulation for extremely challenging cases.
    \item We open-source the \emph{complicated} optical device datatset and code at \href{https://github.com/zhuhanqing/PACE-Light}{PACE-Light} to facilitate AI for PDE community.
\end{itemize}
\section{Preliminaries}
\label{sec:Prelim}
\subsection{Neural Operators for PDE}
Recently, neural operators have emerged as a novel approach for developing machine learning models aimed at solving partial differential equations (PDEs). 
These models focus on learning the mapping between the function spaces in a purely data-driven fashion. 
This holds the generalization capability within a family of PDEs and can potentially be adapted to different discretizations.
Various function bases are utilized to build the operator learning model, such as the Fourier bases~\cite{NN_ICLR2021_Li, NN_ICLR2023_Tran, ML_Arxiv2022_Ashiqur, NP_Neurips2022_Gu}, wavelet bases~\cite{NN_NeurIPS2021_Gupta}, spectral method~\cite{ML_ICML2023_Wu}, and attention layer~\cite{ML_Neurips2024_Li, cao2021choose, li2023transformer}.
These models have demonstrated remarkable performance and efficiency in solving specific types of problems, often achieving record-breaking results in certain applications.
Despite their successes, it's important to recognize that the field of PDEs encompasses a wide variety of equations, each with its own unique properties and characteristics. As pointed out in recent research~\cite{ML_Neurips2024_Li}, there is no guarantee that a single type of data-driven model can effectively address all types of PDEs.

\subsection{Optical Field Simulation with Machine Learning}

Analyzing the propagation of light through optical devices is crucial for the optimization and design of photonic circuits.
For a linear isotropic optical device, with a time-harmonic continuous-wave light beam shining on its input port, we can obtain the steady-state electromagnetic field distributions $\VE(\vec{r})=\normx\Ex+\normy\Ey+\normz\Ez$ and $\VH(\vec{r})=\normx\Hx+\normy\Hy+\normz\Hz$ by solving the steady-state frequency-domain \emph{curl-of-curl} Maxwell PDE under absorptive boundary conditions~\cite{NP_ACSPhotonics2018_Lim}, 
\begin{equation}
    \label{eq:Maxwell}
    \begin{aligned}
        \big((\mu_0^{-1}\curl\curl)-\omega^2\epsilon_0\eps_r(\vec{r})\big)\VE(\vec{r})=j\omega\J_e(\vec{r}),~\big(\curl(\eps_r^{-1}(\vec{r})\curl)-\omega^2\mu_0\epsilon_0\big)\VH(\vec{r})=j\omega \J_m(\vec{r})
    \end{aligned}
\end{equation}
where $\curl$ is the curl operator, $\mu_0$ is the vacuum magnetic permeability, $\epsilon_0$ is the vacuum electric permittivity, $\eps_r$ is the relative electric permittivity, and $\J_m$ and $\J_e$ are the magnetic and electric current sources, respectively.
The finite difference frequency domain (FDFD) method, a widely adopted numerical technique detailed in~\cite{NP_ACSPhotonics2018_Lim}, is used to discretize these continuous-domain equations into an $M\times N$ mesh grid. This transforms the Maxwell PDEs into a linear system $\mathbf{AX}=\vec{b}$. Solving this system with a large sparse matrix $\mathbf{A}\in\mathbb{C}^{MN\times MN}$ is computationally expensive and challenging to scale. Although improvements have been made, such as replacing the \texttt{scipy} solver with the more efficient \texttt{pardiso} solver, the process remains prohibitively costly for large-scale applications.

Building neural networks (NNs) to accelerate this time-consuming simulation process has been investigated in predicting some key design parameters~\cite{NP_SciRep2019_Tahersima, gu2024m3icro} or the entire optical field~\cite{NP_SciRep2019_Trivedi, NP_APL2022_Lim, NP_Nature2021_Chen, NP_Neurips2022_Gu}.
\neurolight extends the neural operator to optical field simulation, enabling learning a physics-agnostic parametric Maxwell PDE solver and achieving SOTA accuracy, while its performance on real-world \emph{complicated} photonic device is still not satisfactory.

\section{Understand the Problem Setup and Challenge}
\label{sec:background}

In this study, we aim to build a physics-agnostic neural operator $\Psi_{\theta}$ for parametric photonic device simulation in a data-driven fashion to approximate the ground-truth Maxwell PDE solver $\Psi^{\ast}:\mathcal{A} \rightarrow \mathcal{U}$ described in Eq.~\eqref{eq:Maxwell}.
Here, $\mathcal{U}$ represents the solution space for the optical field in $\mathbb{C}^{\Omega\times d_u}$ and
$\mathcal{A}=(\Omega, \eps_r, \omega, \mathbf{J})$ represents the observation space of the Maxwell PDE, both defined over the continuous 2-D physical solving domain $\Omega=(l_x,l_z)$.
We follow \neurolight~\cite{NP_Neurips2022_Gu} to discretize the simulation domain $\Omega$ as $\widetilde{\Omega}=(M, N, \Delta l_x, \Delta l_z)$ with adaptive mesh granularity, i.e., with grid steps $\Delta l_x=l_x/M$ and $\Delta l_z=l_z/N$. 
Moreover, 
($\widetilde{\Omega}, \eps_r, \omega$) in the raw observation $\mathcal{A}$ is encoded as informative wave priors, $\mathcal{P}_z=e^{j\frac{2\pi\sqrt{\eps_r}}{\lambda}\mathbf{1}z^{T}\Delta l_z}$ and $\mathcal{P}_x=e^{j\frac{2\pi\sqrt{\eps_r}}{\lambda}x\mathbf{1}^{T}\Delta l_x}$, where $x=(0,1,\cdots,M-1)$ and $z=(0,1,\cdots,N-1)$, reflecting the propagation behaviors of light through different media.
The input light source $\mathbf{J}$ is further modeled as a masked light source field $\VH_y^J$.

Therefore, as illustrated in Fig.~\ref{fig:Teaser}(e), the overarching objective is formulated as learning operator $\Psi_{\theta}$ that maps $\mathcal{A}^{\dagger}=(\eps_r,\VH_y^J,\mathcal{P}_x,\mathcal{P}_z)$ to the target field $\mathcal{U}$ by optimizing the empirical error,
\begin{equation}
    \label{eq:Formulation}
    \theta^{\ast}=\min_{\theta}\mathbb{E}_{a\sim\mathcal{A}^{\dagger}}\big[\mathcal{L}\big(\Psi_{\theta}(a), u\big)\big],
\end{equation}

\subsection{Challenges in Predicting the Light Field of \textit{Complicated} Photonic Devices}
\label{subsec: challenge}

\neurolight~\cite{NP_Neurips2022_Gu} delivers a pioneering effort in extending neural operators to the simulation of photonic devices, achieving SOTA accuracy.
However, it still yields significant errors, particularly for real-world 
complicated devices, with a reported 0.38 normalized mean absolute error for etched MMI device~\cite{NP_SciRep2019_Tahersima, NP_OE2023_Kim}.
This leads us to an interesting reflection: despite the successes of neural operators in solving scientific PDEs, why do they still fall short in \emph{complicated} photonic device simulation?
Below, we provide a detailed analysis that highlights the underlying learning challenges.
\begin{enumerate}[align=left, leftmargin=*]
    \item[\ding{202}] 
    \textbf{Complicated light-matter interaction in the optical field of real-world photonic device}.
    Permittivity $\eps_r$, a critical parameter in photonic devices, greatly impacts how light propagates through media. 
    Designing new devices often involves manipulating the $\eps_r$ distribution across the domain. 
    However, due to manufacturing limitations, $\eps_r$ changes are discrete rather than smooth.
    Moreover, researchers explore patterning materials with highly contrast permittivity to design compact devices~\cite{NP_SciRep2019_Tahersima, verfurth2019heterogeneous}.
    This discrete and highly contrasting permittivity transforms the Maxwell PDE into a \textit{multiscale PDE problem}~\cite{ammari2012multi}, with complicated light-matter interactions such as scattering resonance happening, shown in Fig.~\ref{fig:Teaser} (a), which has been shown difficult to predict from both scientific computing and operator learning perspectives~\cite{NP_ohlberger2018new, xu2023dilated}. 
    \item[\ding{203}] \textbf{Significant prediction field variations from minor structural changes}.
    Due to the complex light-matter interactions within the field, even a slight change in the photonic structure can result in drastically different optical fields under the same input conditions, as shown in Fig.~\ref{fig:Teaser}(b). 
    This calls for a powerful backbone model that is capable of building the relationship between local rival changes with the global optical field transition.
    \item[\ding{204}] \textbf{Non-uniform learning difficulty along the spatial domain}.\label{chal:biased_learning}
    As shown in Fig.~\ref{fig:Teaser}(c),
    with light shining in from a specific position and direction, it propagates through the media, resulting in non-uniform learning difficulties along the spatial domain. 
    Due to the vast diversity of potential internal structures along the light propagation path,
    the light patterns are becoming highly diverse.
    Consequently, the data collected for training also incorporates the same phenomenon where many similar patterns are seen during training near the input sources, whereas the model faces more diverse patterns at greater distances.
    This makes it hard for the model to learn how to predict further regions, especially when the domain is elongated. 
    This issue is analogous to the roll-out error encountered in temporal PDE modeling at the large time steps.
    \item[\ding{205}] \textbf{Rich frequency information lies in the predicted field}.
    We show the energy spectrum of the optical field in the frequency domain in Fig.~\ref{fig:Teaser} (d).
    The field, characterized by complex interactions such as scattering and resonance, exhibits rich frequency information, unveiling the learning complexity from a frequency-domain analysis.
    This confirms the usage of high-frequency modes in \neurolight, underscoring the need for a parameter-efficient, robust, and powerful backbone model to resolve the parameter efficiency and overfitting issue with large modes.
\end{enumerate}

\section{Proposed \name Methods}
\label{sec:method}
In this paper, we follow the standard operator learning model architecture as 
\begin{equation}
    \label{eq:arch}
    a^{\dagger}({\vec{r}}) \rightarrow v_0(\vec{r}) \rightarrow v_1(\vec{r}) \rightarrow \cdots v_K(\vec{r}) \rightarrow u(\vec{r}), ~~~ \forall \vec{r} \in \Omega.
\end{equation}
We start with the convolutional stem used in~\cite{NP_Neurips2022_Gu} to project the PDE observation $a^{\dagger}(\vec{r})$ into a higher-dimensional feature space of dimension $C$.
This is followed by a sequence of $K$ cascaded neural operator blocks, which gradually reconstruct the complex optical field within the $C$ dimensional space. 
At last, a head with two point-wise convolutional layers projects the $v_{K}(\vec{r})$ to the optical field space $u(\vec{r})$.
Fig.~\ref{fig:model}(a) shows the proposed \name neural operator block structure, formulated as,
\begin{equation}
    \label{eq:PaceBlock}
    v_{k+1}(\vec{r}):= \texttt{FFN}\big((\mathcal{K}v_{k}^{'})(\vec{r}) + v_k\big) + v_k,~\forall \vec{r}\in\Omega;~ v_k^{'} (\vec{r}) = \text{pre-norm} (v_k(\vec{r})),
\end{equation}
where $\mathcal{K}$ is the our proposed \name operator and $\text{FFN} (\cdot)$ is a feedword network used in~\cite{NP_Neurips2022_Gu}.
To stabilize the model performance when scaling to deeper layers, we add pre-normalization~\cite{xiong2020layer} and follow~\cite{TensorizedFNO} to add a double skip.
In this work, we consistently use the \neurolight operator in the first two blocks to align our model with the horizontal and vertical wave prior encoding method adopted from \neurolight, which we found slightly improves our accuracy.

\subsection{Parameter-efficient and Effective Cross-axis Factorized \name Operator}
\label{subsec:backbone}

\begin{figure}[t]
    \centering
\includegraphics[width=0.95\textwidth]{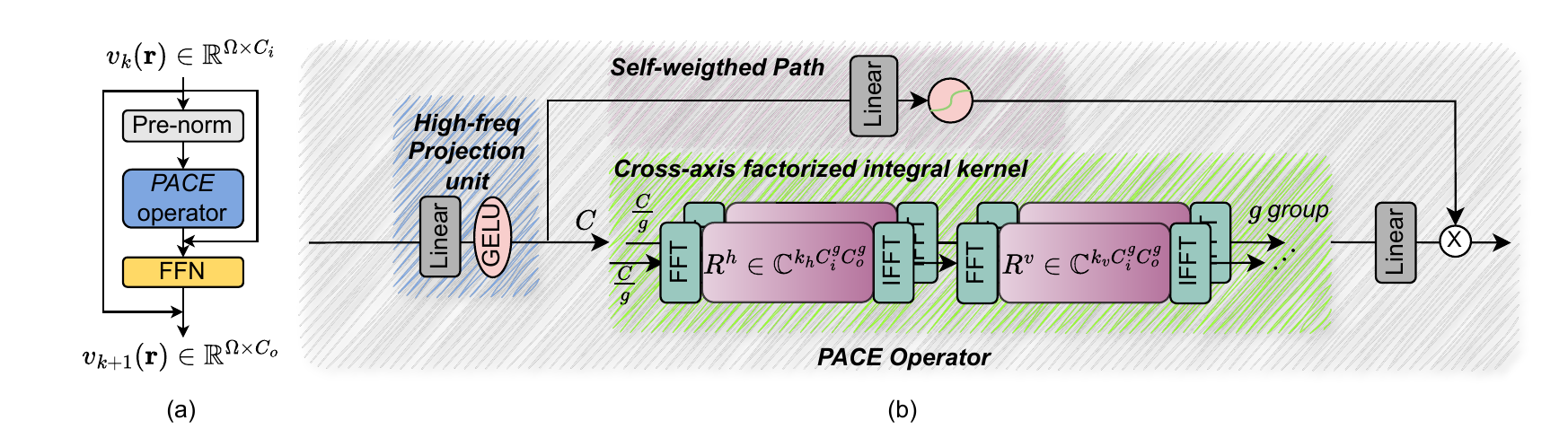}
    \caption{\small (a) \name block with double skip and pre-normalization;
    (b) Our cross-axis factorized \name operator.}
    \label{fig:model}
    \vspace{-15pt}
\end{figure}

The neural operator design is key to obtaining satisfactory accuracy on a given PDE task. 
With the well-discussed challenges in Sec.~\ref{subsec: challenge}, we derive key insights that have guided the development of our \name operator in Fig.~\ref{fig:model}(b): (1) Long-distance full-domain modeling capacity, especially effectively modeling how local features impact the whole domain;
(2) Isotropic model architecture with no down-sampling/ patching without losing local details;
(3) Parameter efficiency under the needs of capturing high-frequency features.

Given the isotropic requirements, an operator based on Fourier bases is an ideal candidate as it achieves full-domain attention in the $O(n log n)$ time complexity.
However, the rich frequency
\begin{wrapfigure}{r}{0.43\textwidth}
\includegraphics[width=\linewidth]{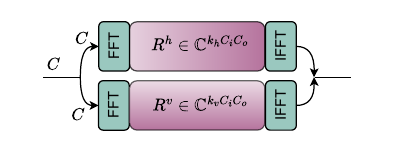} 
\caption{\small Factorized FNO~\cite{ML_ICLR2023_Tran, NP_Neurips2022_Gu}.}
\label{fig:ffno}
\vspace{-5pt}
\end{wrapfigure}
information lying in the optical field requires the use of large frequency modes, making the FNO~\cite{NN_ICLR2021_Li} with huge parameters and severe overfitting issues.
\neurolight~\cite{NP_Neurips2022_Gu} and Factorized FNO~\cite{NN_ICLR2023_Tran} propose to decompose the FNO block with independent 1-D FNO blocks in the full $N$-dimensional domain $\Omega$ (see Fig.~\ref{fig:ffno}), therefore, solving the parameter concern when utilizing high-frequency modes and serving as a regularization for overfitting.
The only difference between \neurolight and Factorized FNO~\cite{NN_ICLR2023_Tran} is whether they chunk the input or copy the input to the independent 1-D FNO block.
We argue that their theoretical success is attributed to the \emph{implicit full-domain integration} in Corollary~\ref{col:single_axis_fact}.

\begin{corollary}\label{col:single_axis_fact}
    The factorized Fourier integral operator $\mathcal{K}$~\cite{NN_ICLR2023_Tran, NP_Neurips2022_Gu} factorizes the original Fourier integral operator~\cite{NN_ICLR2021_Li} along each dimension $n$ in the N-dimension domain $\Omega$,
    \begin{equation}
    \label{eq:factfno}
        \begin{aligned}
            (\mathcal{K}v_k)(\vec{r_1})=\sum_{n}^{N}\mathcal{F}^{-1}_{n}(\mathcal{F}_{n} (\kappa_{\phi}^{n}) \cdot \mathcal{F}_{n}(\vec{r}_2)) (\vec{r}_1), ~~~ \forall \vec{r}_1\in \Omega,
        \end{aligned}
    \end{equation}
    where each item explicitly computes a 1-D kernel integral, $\int_{\Omega_n}\kappa(\vec{r}_1, \vec{r}_2)^{n}v_k(\vec{r}_2)^{n}\text{d}v_k(\vec{r}_2)^{n}$.
    It implicitly implements full-domain kernel integration in $\Omega$ by stacking $\mathcal{K}$, i.e., $\mathcal{K}_0 \circ \mathcal{K}_1 \circ \cdots$, 
\end{corollary}

However, the reliance on implementing full-domain integration with multi-layers makes them \emph{weak} operator candidates to achieve our first requirement, i.e., a \emph{strong} model that is capable of building \textit{full-domain} modeling between local structures with the global fields.

\noindent\textbf{Proposed cross-axis 2-D factorized integral kernel}.
Aware of the above shortcomings of previous factorized FNO variants, in our 2-D domain, we propose to factorize the full domain integral in a cross-axis way along the horizontal (h) and vertical (v) axis: 
\begin{equation}
    \label{eq:crossintegral}
    \begin{aligned}
        (\mathcal{K}v_k)(\vec{r_1}) &=\int_{\Omega}\kappa(\vec{r}_1, \vec{r}_2)v_k(\vec{r}_2)\text{d}v_k(\vec{r}_2), ~~~ \forall \vec{r}_1\in \Omega, \\
        &\approx \int_{\Omega_h}\kappa(\vec{r}_1, \vec{r}_2)^h\int_{\Omega_h} \kappa(\vec{r}_1, \vec{r}_2)^v v_k(\vec{r}_2)\text{d}v_k(\vec{r}_2)^v \text{d}v_k(\vec{r}_2)^h, ~~~ \forall \vec{r}_1\in \Omega.
    \end{aligned}
\end{equation}
This factorization enables an \emph{explicit factorized full-domain integration}. It provides a \emph{strong} way to capture the relationship between points in the domain $\Omega$, building the relationship between local structure with the complicated field pattern.
The implementation of the above cross-axis integral can be efficiently implemented by Fourier Transform $\mathcal{F}(\cdot)$ when the kernel $\kappa(r_1, r_2) = \kappa(r_1-r_2)$, as follows,
\begin{equation}
\label{eq:cfno}
    \begin{aligned}
        (\mathcal{K}v_k)(\vec{r_1})= \mathcal{F}^{-1}_{h}(\mathcal{F}_{h} (\kappa^{h}) \cdot \mathcal{F}_{h} (\mathcal{F}^{-1}_{z}(\mathcal{F}_{v} (\kappa^{v}) \cdot \mathcal{F}_{z}(\vec{r}_2))) (\vec{r}_1), ~~~ \forall \vec{r}_1\in \Omega,
    \end{aligned}
\end{equation}
in a $nlogn$ complexity ($n=MN$ in our 2-D cases with $\Omega \in \mathbb{C}^{M \times N}$).

\noindent\textbf{Group-wise cross-axis integration}.
For input $\Vec{r}$ with a channel dimension $C$, it can be viewed as the sampling of a set of functions $\{r_l(\cdot, \cdot)\}_{l=1}^{C}$ on grid point in the 2-D discretized domain $\Omega = \Omega_h \times \Omega_v$.
The learnable integral kernel intrinsically performs information exchange along different grid points in $\Omega$.
Similar to the multi-head design in Transformer, which assumes different heads extract different information, we can also partition the $C$ basis functions into $g$ disjoint sub-groups and feed each sub-group through our cross-axis factorized kernel.
This grouping further reduces the number of parameters to $(\kappa_h + \kappa_v) \times \frac{C_o C_i}{g}$, showing significant parameter reduction compared to FNO ($\kappa_h \times \kappa_v\times C_o C_i$) and Factorized FNO ($(\kappa_h + \kappa_v)\times C_o C_i$), showing excellent parameter efficiency when utilizing large frequency modes is a must.
We do an ablation study in Appendix~\ref{subsec:group} to investigate the choices of different group $g$, where we find $g=4$ strikes the best between parameter efficiency and model performance.

\noindent \textbf{Explicit projection unit $\xi$ for extracting high frequency information}.
The optical field shows rich information in the frequency spectrum, reciting a special care of high-frequency information.
Besides utilizing high-frequency modes, we propose to add an explicit projection module before the cross-axis integral, which is very simple as one linear layer followed by a non-linear activation, given non-linear activation is known to help generate high-frequency features~\cite{NN_Neurips2024_Bogdan}.

\noindent \textbf{Self-weighted path for enhanced instance-based local feature attention}.
The optical field's response is intricately linked to the minute variations in different photonic device structures. 
A self-weighted path is introduced to ensure the model can pay different attention to regions of significant influences for varying device structures. 
An instance-based weight is generated by passing the feature map after the projection unit through a linear layer and a \texttt{Sigmoid} unit, and then multiplied with the results after the cross-axis integral unit to provide instance-based attention.

Overall, the above ingredients are assembled together as our proposed \name operator, as shown in Fig.~\ref{fig:model} (b), which implements a self-weighted 2-D cross-axis factorized integral transform.

\subsection{Cascaded Learning from Rough to Clear}
\begin{figure}
    \centering
    \includegraphics[width=0.95\linewidth]{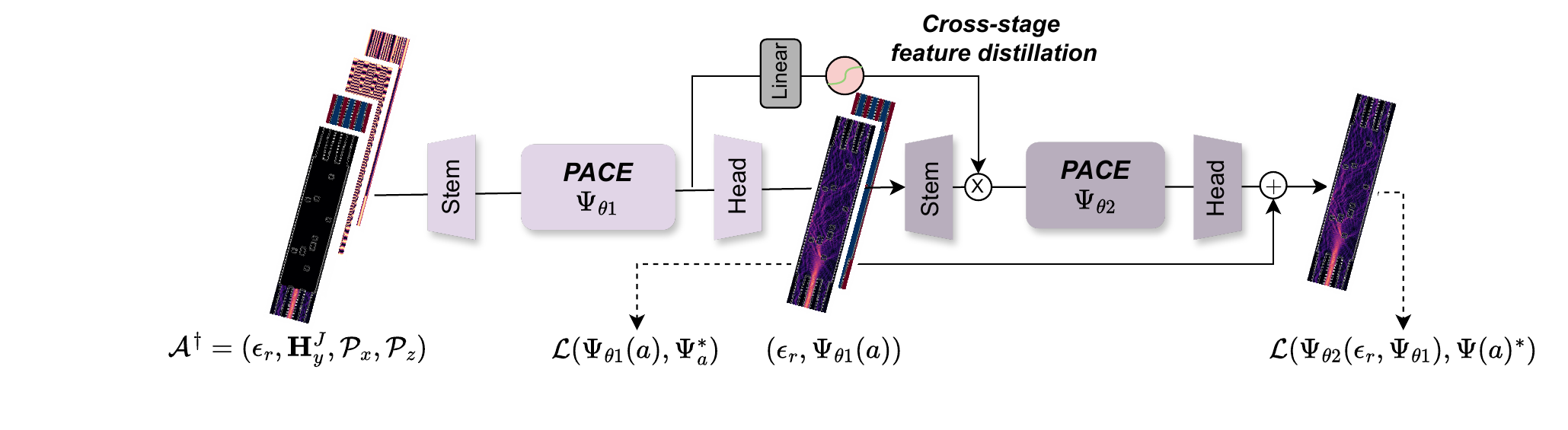}
    \caption{\small The proposed cascaded learning flow with two stages. The first stage learns an initial and rough solution, followed by the second stage to revise it further. A cross-stage distillation path is used to transfer the learned knowledge from the first stage to the second stage.}
    \label{fig:twostage}
    \vspace{-15pt}
\end{figure}

With the effective \name operator design, the prediction fidelity can be largely improved by only using a 12-layer \name model (see Section. \ref{exp_main_1}).
But for some complicated benchmarks (e.g., etched MMI 3x3/5x5), it still yields $\sim$ 10\% mean squared error, which is not satisfying.
A straightforward solution might involve scaling up the model size, expecting additional layers would enhance performance. 
However, as demonstrated in~\cite{ML_ICLR2023_Tran}, scaling to deep layers shows saturated performance after exceeding a specific number.

Existing ML for PDE solving work typically learns a model in a one-shot way by directly learning the underlying relationship from input-output pairs. Unlike AI systems, humans don’t learn new and difficult tasks in a one-shot manner; instead, they learn skills progressively, starting with easier tasks and gradually moving to harder ones. For example, instead of directly learning how to solve equations, students first learn basic operations, such as addition and multiplication, and then move on to solving complex equations.

Hence, inspired by this human learning process, unlike previous work that directly learns a one-stage model, we propose to divide the challenging optical field prediction problem into two sequential latent tasks.
The first task, undergoing the same problem setup as discussed in Sec.~\ref{sec:background}, could predict an initial, rough optical field based on the less informative raw PDE observation (we only have the light source and device permittivity distribution).
Then, the successive second task could refine the rough prediction further by capturing more details and nuances, by accepting the predicted field $\Psi_{\theta_1}$ and device permittivity $\eps_r$ as the input. 
Therefore, we assign \emph{higher Fourier modes} to enable sufficient capacity.
The divide-and-conquer way results in a cascaded two-stage model architecture, as shown in Fig.~\ref{fig:twostage}.
The cascaded learning model is trained jointly (\namea + \nameb) with the optimization target as the sum of two losses $\mathcal{L}(\Psi_{\theta1} (a), u) + \mathcal{L}(\Psi_{\theta2} (\Psi_{\theta1} (a), \eps_r), u)$, where the first $\mathcal{L}(\Psi_{\theta1} (a), u)$ serves as intermediate supervision that enfores the first stage model condensate the learned knowledge.
To better connect the two-stage model, we propose a \emph{cross-stage feature distillation path} to distill learned feature from the previous stage to the last by using a simple \texttt{Linear}$\rightarrow$\texttt{Sigmoid} path.
\section{Experimental Results}
\label{sec:Results}
\subsection{Experimental Setup}

\textbf{Benchmarks:} 
We evaluate our methods on real-world \emph{complicated} photonic devices that pose significant simulation challenges for ML surrogate models. 
This includes the Etched MMI with randomly placed rectangular cavities, used in~\cite{NP_Neurips2022_Gu}, and the metaline device~\cite{zarei2020integrated, meinzer2014plasmonic} featuring two layers of randomly dimensioned meta-atoms. 
\textbf{These devices present a highly discrete and contrast permittivity distribution and complex light-matter interactions, making them ideal for testing the effectiveness of our model. }
We generate our datasets using the open-source 2-D FDFD simulator, Angler \cite{NP_ACSPhotonics2018_Lim}, with generation details in Appendix~\ref{subsec:dataset}. 

\noindent\textbf{Baselines}: We evaluate the proposed \name model against a range of baselines, including the SOTA neural operator work, \neurolight~\cite{NP_Neurips2022_Gu}, for optical simulation. 
We also include representative operator learning models for scientific PDEs based on Fourier bases(FNO~\cite{NN_ICLR2021_Li}, Factorized FNO (F-FNO)~\cite{NN_NeurIPSW2021_Tran, NN_ICLR2023_Tran}, U-NO~\cite{ML_Arxiv2022_Ashiqur}, tensorized FNO (TFNO)~\cite{TensorizedFNO}), 
attention kernels~\cite{ML_Neurips2024_Li}, and the latent spectral method (LSM)~\cite{ML_ICML2023_Wu}. 
We also incorporate UNet~\cite{NP_APL2022_Lim, NP_Nature2021_Chen} and Dilated ResNet (Dil-ResNet) ~\cite{NN_ICLR2021_Kim}.
For a fair comparison, 
we keep a model size budget of under/near 4 million (M)  parameters for baselines, except LSM~\cite{ML_ICML2023_Wu} where the original implementation is adopted.
Details on model configurations are in the Appendix~\ref{subsec:model_details}.

\noindent\textbf{Training setting and metric}:
All models undergo training for 100 epochs using the AdamW optimizer with a weight decay of $1e^{-5}$ in a batch size of 4.
To balance the optimization among different fields,
we use normalized mean squared error (N-MSE) as the learning objective,
\begin{equation}
\begin{aligned}
    &\mathcal{L}\big(\Psi_{\theta}(a),\Psi^{\ast}(a)\big)=(\|\Psi_{\theta}(\mathcal{E}(a))-\Psi^{\ast}(a)\|^2)/\|\Psi^{\ast}(a)\|^2.
\end{aligned}
\end{equation}
We don't use the previously-used mean absolute error (MAE)~\cite{NP_Neurips2022_Gu} as the metric given for complex-valued optical fields; we argue that L2 distance is a more accurate metric to evaluate the distance in the complex plane with a detailed analysis in Appendix~\ref{subsec:mse}.
We adopt the superposition-based mix-up technique~\cite{NP_Neurips2022_Gu} to generate input light combinations randomly to augment training data.

\subsection{Main Results}
\begin{table}
    \centering
    \captionof{table}{\small \small Comparison of \# parameters, training error (last epoch), and test error on three benchmarks among our \name and various baselines. We use geo-means to report overall improvements across different benchmarks.}
    \label{tab:MainResults}
    \resizebox{0.99\textwidth}{!}{%
    \begin{tabular}{cc|ccc}
    \toprule
    Benchmarks                    & Model                              & \multicolumn{1}{l}{\#Params (M) $\downarrow$} & \multicolumn{1}{c}{Train Err ($10^{-2}$) $\downarrow$}                         & \multicolumn{1}{c}{Test Err ($10^{-2}$) $\downarrow$}                          \\ \midrule
                                  & UNet~\cite{NP_APL2022_Lim,NP_Nature2021_Chen}                                        & 3.88                                  & 63.03                                 & 65.32                                  \\
                                  & Dil-ResNet~\cite{NN_ICLR2021_Kim}                                        & 4.17                                  & 51.34                                  & 51.79                                 \\
                                  & Attention-based model~\cite{ML_Neurips2024_Li}                                        & 3.75                                  & 70.05                                 & 69.85                                  \\
                                  & U-NO~\cite{ML_Arxiv2022_Ashiqur}                                        & 4.38                                  & 34.22                                 & 42.86                                  \\
                                  & Latent-spectral method~\cite{ML_ICML2023_Wu}                                        & 4.81                                  & 55.07                                 & 55.16                                  \\
                                  & FNO-2d~\cite{NN_ICLR2021_Li}                                      & 3.99                                  & 32.51                                  & 38.71                                 \\
                                  & Tensorized FNO-2d~\cite{TensorizedFNO}                                      & 2.25                                  & 35.52                                  & 36.61                                 \\
                                  & Factorized FNO-2d ~\cite{NN_ICLR2023_Tran}                                       & 4.02                                  & 24.2                                 & 32.81                                   \\
                                  & \neurolight ~\cite{NP_Neurips2022_Gu}                                       & 2.11                                  & 15.58                                  & 17.21                                  \\
     \multirow{-10}{*}{\begin{tabular}{c}Etched MMI 3x3 \\ \includegraphics[width=4cm]{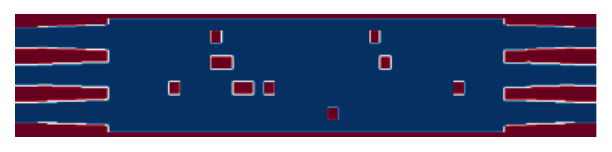} \\
     \includegraphics[width=4cm]{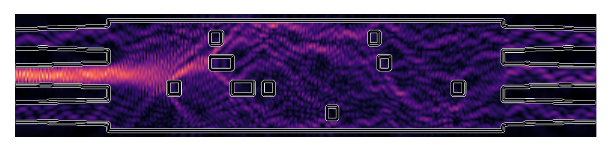} \end{tabular}} & \cellcolor[HTML]{CCEECC}\textbf{\name} & \cellcolor[HTML]{CCEECC}\textbf{1.71} & \cellcolor[HTML]{CCEECC}\textbf{9.51} & \cellcolor[HTML]{CCEECC}\textbf{10.59} \\ \midrule
    & UNet~\cite{NP_APL2022_Lim,NP_Nature2021_Chen}                                        & 3.88                                  & 65.73                                  & 66.01                                 \\
                                  & Attention-based model~\cite{ML_Neurips2024_Li}                                        & 3.75                                  & 74.16                                 & 74.20                                  \\
                                  & U-NO~\cite{ML_Arxiv2022_Ashiqur}                                        & 4.38                                  & 37.92                                  & 42.24                                  \\
                                  & Latent-spectral method~\cite{ML_ICML2023_Wu}                                        & 4.81                                  & 53.9                                 & 54.01                                  \\
                                  & FNO-2d~\cite{NN_ICLR2021_Li}                                      & 3.99                                  & 33.12                                  & 36.49                                 \\
                                  & Tensorized FNO-2d~\cite{TensorizedFNO}                                      & 2.25                                  & 39.11                                  & 39.45                                 \\
                                  & Factorized FNO-2d ~\cite{NN_ICLR2023_Tran}                                       & 4.02                                  & 22.18                                  & 26.06                                  \\
                                  & \neurolight ~\cite{NN_NeurIPSW2021_Tran}                                       & 2.11                                  & 18.04                                  & 17.41                                  \\
    \multirow{-10}{*}{\begin{tabular}{c}Etched MMI 5x5 \\ \includegraphics[width=4cm]{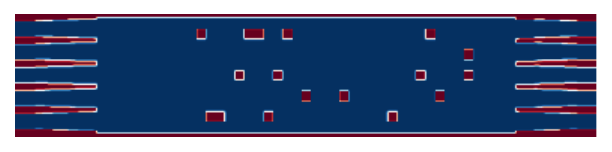} \\
    \includegraphics[width=4cm]{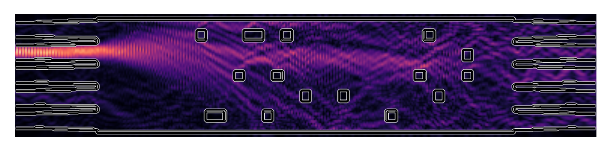} \end{tabular}} & \cellcolor[HTML]{CCEECC}\textbf{\name} & \cellcolor[HTML]{CCEECC}\textbf{1.71} & \cellcolor[HTML]{CCEECC}\textbf{11.66} & \cellcolor[HTML]{CCEECC}\textbf{11.91} \\ \midrule
    & UNet~\cite{NP_APL2022_Lim,NP_Nature2021_Chen}                                        & 3.88                                  & 39.12                                 & 39.61                                  \\
                                  & Dil-ResNet~\cite{NN_ICLR2021_Kim}                                        & 4.17                                  & 12.37                                  & 13.20                                 \\
                                  & Attention-based model~\cite{ML_Neurips2024_Li}                                        & 3.75                                  & 63.99                                & 64.10                                 \\
                                  & U-NO~\cite{ML_Arxiv2022_Ashiqur}                                        & 4.38                                  & 19.27                                 & 22.09                                  \\
                                  & Latent-spectral method~\cite{ML_ICML2023_Wu}                                        & 4.81                                  & 31.60                                 & 31.94                                  \\
                                  & FNO-2d~\cite{NN_ICLR2021_Li}                                      & 3.21                                  & 19.73                                  & 20.88                                 \\
                                  & Tensorized FNO-2d~\cite{TensorizedFNO}                                      & 1.58                                  & 30.60                                  & 31.04                                 \\
                                  & Factorized FNO-2d ~\cite{NN_ICLR2023_Tran}                                       & 2.68                                  & 8.51                                  & 9.28                                  \\
                                  & \neurolight ~\cite{NP_Neurips2022_Gu}                                       & 1.49                                  & 6.76                                  & 6.09                                  \\
     \multirow{-10}{*}{\begin{tabular}{cc} \multicolumn{2}{c}{Metaline 3x3} \\ \includegraphics[width=2cm]{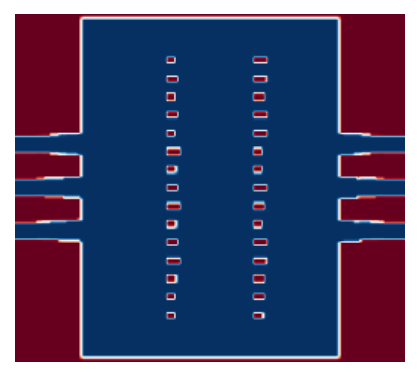} &
     \includegraphics[width=2cm]{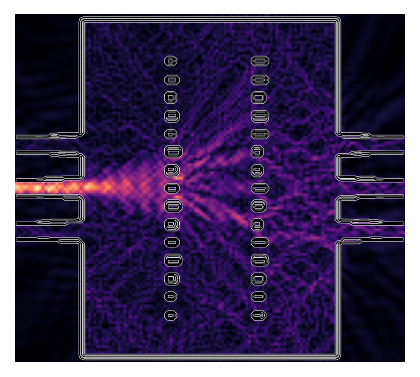}
     \end{tabular}} & \cellcolor[HTML]{CCEECC}\textbf{\name} & \cellcolor[HTML]{CCEECC}\textbf{1.24} & \cellcolor[HTML]{CCEECC}\textbf{3.32} & \cellcolor[HTML]{CCEECC}\textbf{2.82} \\ \midrule
    \multicolumn{2}{c|}{Improvement over best baseline \neurolight ~\cite{NP_Neurips2022_Gu} }                           & \cellcolor[HTML]{CCEECC}\textbf{-17.70\%}                 & \cellcolor[HTML]{CCEECC}\textbf{-41.23\%}              & \cellcolor[HTML]{CCEECC}\textbf{-39.03\%} \\
    \multicolumn{2}{c|}{Improvement over all baselines}                           & \cellcolor[HTML]{CCEECC}\textbf{-51.67\%}                 & \cellcolor[HTML]{CCEECC}\textbf{-72.57\%}              & \cellcolor[HTML]{CCEECC}\textbf{-73.85\%} 
    \\\bottomrule
    \end{tabular}%
    }
    \vspace{-10pt}
\end{table}
\subsubsection{Prediction Quality of Single \name Model}\label{exp_main_1}
In Tab.~\ref{tab:MainResults}, we compare our 12-layer \name model with various baselines on multiple real-world device benchmarks, showing significant \textbf{73.85\%} smaller test error with \textbf{51.67\%} fewer parameters on average.
Notably, even when compared to the \emph{best} baseline, 16-layer \neurolight, we show over \textbf{39\%} lower test error with over \textbf{17\%} fewer parameters.
Given the challenge \ding{203} that \textit{trial structure change can totally change the optical field},
model relying on downsampling or patching fails to capture the local details, confirming the failure of the UNet and Transformer model.
Moreover, the challenge \ding{202} and challenge \ding{204} call for a powerful model with long-distance modeling capability.
Although Dil-ResNet utilizes a dilated block to enlarge the receptive field, it is insufficient for a large domain, validated by the result that it shows much better accuracy on the small Metaline than the etched MMI3x3.
Capturing long-range dependency with the Fourier operator provides an efficient way to the isotropic model without any downsampling, therefore making the Fourier-operator type model show consistently better accuracy than other baseline methods.
However, due to the challenge \ding{205} that there is rich frequency information in the predicted field, FNO-2d falls short due to the impediment of utilizing large modes given the large parameter count.
We also compared it with the tensorized FNO 2d. However, we find the general tensor decomposition hurt the accuracy of this challenging task.
\neurolight shares a similar insight of Factorized FNO by factoring Fourier kernel with several independent 1-D Fourier kernels; however, as we argued before, it fails to establish a strong full-domain modeling capacity by linking local details to the global complex field.
Overall, our \name block benefits from a physically meaningful cross-axis Fourier kernel factorization, equipping the capacity to capture full-domain dependency in a parameter-efficient way.
Visualization of predicted results is in Appendix~\ref{subsec: visual}.

\subsubsection{Quality Improvement with Two-stage Model}
\begin{table}
    \centering
    \captionof{table}{\small Comparison between our two-stage model and simply scaling more layers. All models use the same Fourier modes setup.}
    \label{tab:2stageResults}
    \resizebox{0.99\textwidth}{!}{%
    \begin{tabular}{ccc|ccc}
    \toprule
    Benchmarks                    & Model      & Cross-stage dist.        & \multicolumn{1}{l}{\#Params (M) $\downarrow$} & \multicolumn{1}{c}{Train Err ($10^{-2}$) $\downarrow$}                         & \multicolumn{1}{c}{Test Err ($10^{-2}$) $\downarrow$}                          \\ \midrule
                                  & \cellcolor[HTML]{C0C0C0}\name-12 layer             &    -                          & \cellcolor[HTML]{C0C0C0}1.73                                 & \cellcolor[HTML]{C0C0C0}9.51                                 & \cellcolor[HTML]{C0C0C0} 10.59                                  \\ \cmidrule{2-6}
                                  & \name-20 layer                    & -                 & 3.135                                  & 6.46                                 & 7.04                                  \\
                                  & \namea $+$ \nameb          & \ding{56}                             & 3.151                                  & 4.66                                 & 5.83                                  \\
                                  \multirow{-3}{*}{Etched MMI 3x3} &  \cellcolor[HTML]{CCEECC}\namea $+$ \nameb          &  \cellcolor[HTML]{CCEECC}\ding{52}                             &  \cellcolor[HTML]{CCEECC}3.151                                  &  \cellcolor[HTML]{CCEECC}4.14                                 &  \cellcolor[HTML]{CCEECC}5.32                                  \\
     \midrule
    & \cellcolor[HTML]{C0C0C0}\name-12 layer             &    -                          & \cellcolor[HTML]{C0C0C0}1.73                                 & \cellcolor[HTML]{C0C0C0} 11.66                                & \cellcolor[HTML]{C0C0C0} 11.91                                 \\ \cmidrule{2-6}
                                  & \name-20layer                    & -                 & 3.135                                  & 7.74                                & 7.88                                 \\
                                  & \namea $+$ \nameb          & \ding{56}                             & 3.151                                  & 6.17                                 & 6.78                                  \\
                                 \multirow{-3}{*}{Etched MMI 5x5} &  \cellcolor[HTML]{CCEECC}\namea $+$ \nameb          &  \cellcolor[HTML]{CCEECC}\ding{52}                             &  \cellcolor[HTML]{CCEECC}3.151                                  &  \cellcolor[HTML]{CCEECC}5.43                                &  \cellcolor[HTML]{CCEECC}6.15                                  \\  
    \bottomrule
    \end{tabular}%
    }
    \vspace{-15pt}
\end{table}
We further compare the proposed cascaded two-stage model with the common practice of solely increasing \# layers.
We set the \namea as a 12-layer \name model with Fourier modes(\#Mode =70, \#Mode =40), and \nameb as a 8-layer \name model with larger Fourier modes (\#Mode =100, \#Mode =40).
As shown in Tab.~\ref{tab:2stageResults}, 
the two-stage setup introduces slight overhead for one extra set of stem and head but shows a clear margin over only increasing the number of layers in terms of both train error and test error.
The cross-stage feature distillation further provides meaningful guidance by transferring learned features to the second-stage model, leading to the best accuracy for the two-stage setup.
In Appendix~\ref{subsec:seq}, we also show that the cross-stage distillation trick can improve model accuracy, similar to a more costly training setup, by training the two-stage models sequentially.

\subsubsection{Speedup over Numerical Tools}

To develop a fast surrogate ML model that can replace the Maxwell PDE solver, it's crucial to
\begin{wrapfigure}{r}{0.4\textwidth}
\vspace{-10pt}
\includegraphics[width=\linewidth]{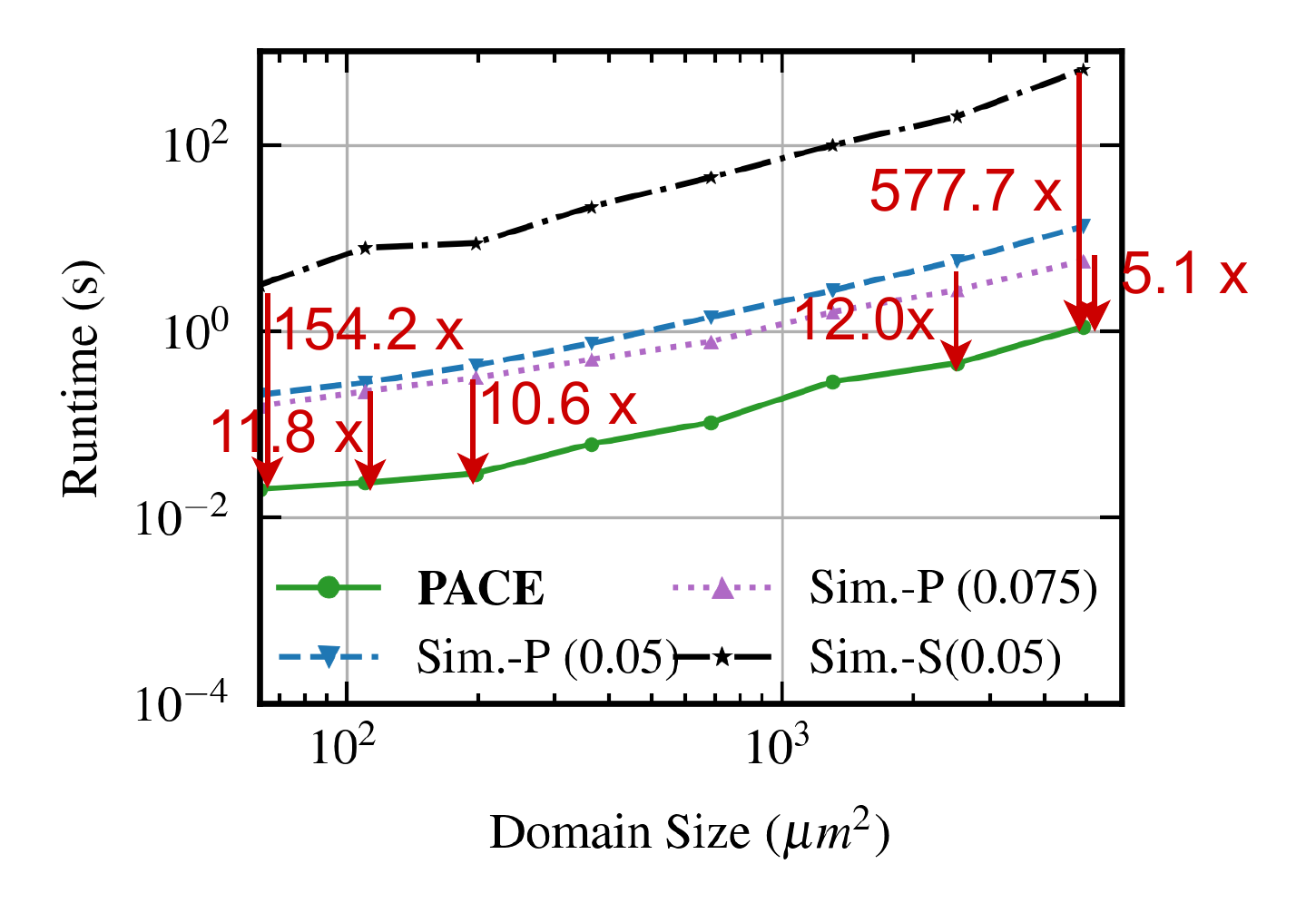} 
\caption{\small Speedup of \name over angler~\cite{NP_ACSPhotonics2018_Lim} using \texttt{scipy} (S)/ \texttt{pardiso} (P) with simulation granularity (0.05nm) and (0.075nm).}
\label{fig:runtime}
\vspace{-5pt}
\end{wrapfigure}
evaluate the speed-up of our \name model compared to the FDFD numerical simulator Angler~\cite{NP_ACSPhotonics2018_Lim}.
We vary the simulation domain size and set the grid step to 0.05 nm, scaling the discretized size \texttt{pardiso} linear solvers, respectively and number of frequency modes to ensure the model has sufficient capacity to capture the entire simulation domain. For comparison, we employ a 20-layer joint \name model. As shown in Fig.~\ref{fig:runtime}, our \name model achieves a speed-up of 150-577$\times$ and 12$\times$ over Angler on a 20-core Intel i7-12700 CPU using the \texttt{scipy} and 
We further set a larger simulation granularity, 0.075 nm, to check speedup if we tolerate simulation quality loss in commercial tools. However, we find that setting a larger granularity results in a significantly different field, as qualitatively shown in reb-Fig.3, with a corresponding N-MSE error of 1.2.
Even though in this case, \name still shows a 5.1-10.6$\times$ speedup over \texttt{pardiso}-based Angler with much better fidelity.

\subsection{Discussion}

\noindent\textbf{Cross-axis \name block design choices}.
\noindent In Tab.~\ref{tab:ModelAblation}, we \emph{independently} alter individual components
\begin{wraptable}{r}{8cm}
\centering
\vspace{-10pt}
\caption{\small~Model design ablation on Metaline dataset.}
\label{tab:ModelAblation}
\resizebox{\linewidth}{!}{%
\begin{tabular}{@{}cccc@{}}
\toprule
Variants & \begin{tabular}[c]{@{}c@{}}\#Params \\ (M)$\downarrow$\end{tabular}  & \begin{tabular}[c]{@{}c@{}}\#Train Err \\ ($10^{-2}$)$\downarrow$\end{tabular}  & \begin{tabular}[c]{@{}c@{}}\#Test Err \\ ($10^{-2}$)$\downarrow$\end{tabular}  \\ \midrule
\textbf{8-layer \name} & \textbf{0.82} & \textbf{5.65} & \textbf{4.82} \\ \midrule
No self-weighted path & 0.8 & 6.33 & 5.66 \textcolor{red}{(+0.84)} \\
No projection unit & 0.8 & 6.58 & 5.97 \textcolor{red}{(+1.15)} \\ 
Use TFNO & 1.06 & 10.80 & 9.51 \textcolor{red}{(+4.69)} \\ 
\bottomrule
\end{tabular}
}
\vspace{-5pt}
\end{wraptable} 
within the \name operator to assess their effectiveness.
The self-weighted path, which provides instance-specific weights, significantly improves model accuracy across various photonic device patterns. Removing this component results in a 17\% increase in error, highlighting its importance. 
Similarly, eliminating the high-frequency projection unit leads to a 23\% worse error, emphasizing its crucial role in capturing high-frequency features. 
To further illustrate this, we visualize the feature maps in the frequency domain before and after applying the nonlinear activation in the high-frequency projection unit. As shown in Fig.~\ref{fig:high_freq}, the nonlinear activation effectively amplifies high-frequency components, supporting our claim and validating the design decision to incorporate an additional high-frequency projection path.
Lastly, we replace our cross-axis Factorized integral kernel with a recent tensorized FNO (TFNO)~\cite{TensorizedFNO} (\texttt{tucker} decomposition with rank 0.02).
While TFNO effectively models long-range dependencies, matching our parameter count required aggressive decomposition, which significantly degraded performance. This comparison underscores the advantage of our physically grounded \textit{cross-axis factorized kernel}.

\noindent\textbf{Generalization to out-of-distribution testing}.
As an operator model that is parameter-agnostic, it 
\begin{wrapfigure}{r}{0.5\textwidth}
\vspace{-10pt}
\includegraphics[width=\linewidth]{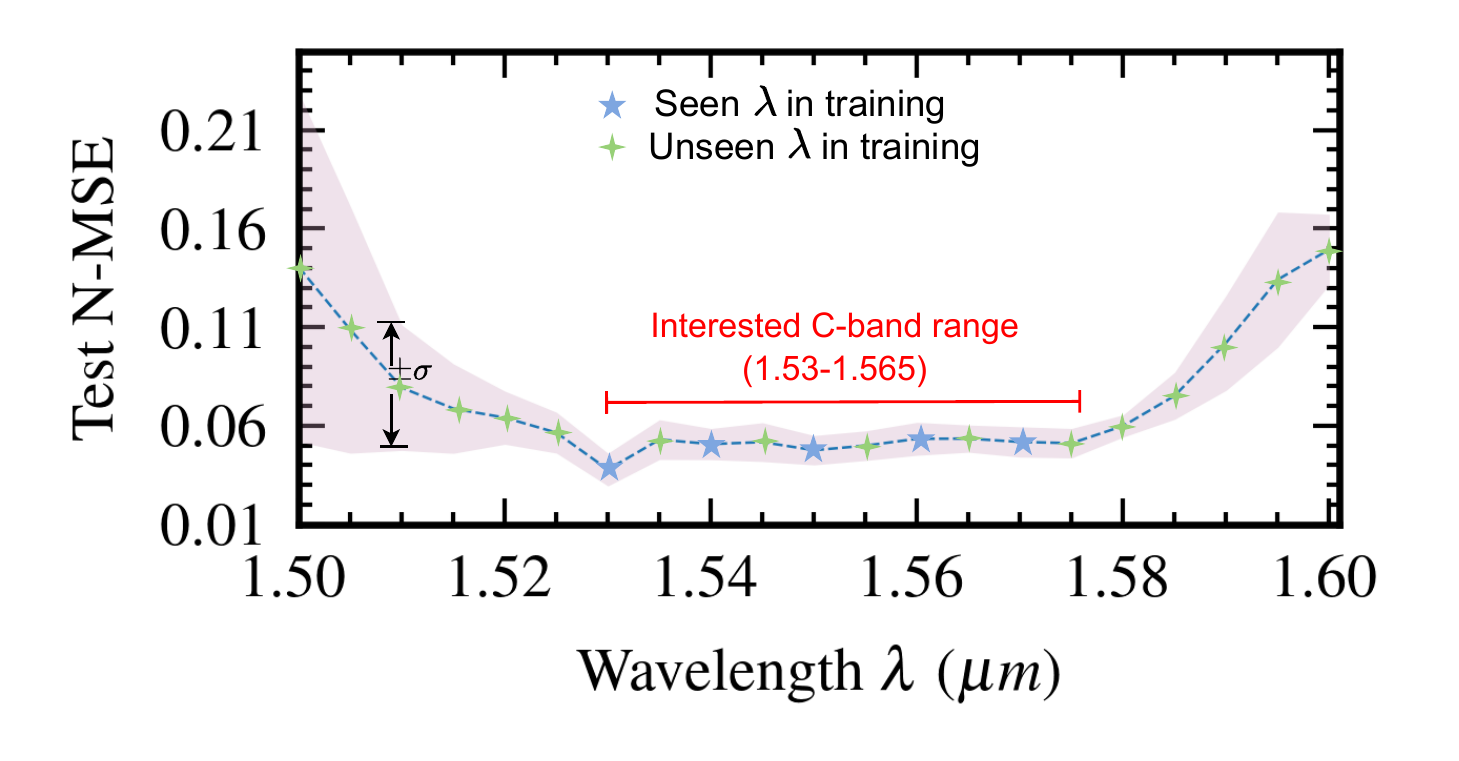} 
\caption{\small Generalize to unseen wavelength in interested C-band (1.53-1.565) and outside C-band.}
\label{fig:genSpec}
\vspace{-5pt}
\end{wrapfigure}
is important to test the generalization for out-of-distribution data with unseen parameters. 
We re-generate photonic devices with different device configurations (size, etched region, etc.) and unseen frequencies in our interested wavelength range (1.53-1.565 $\mu$m), i.e., C-band.
As shown in Fig.~\ref{fig:genSpec}, our \name model generalizes well on unseen simulation frequency and new devices.
It is a vital test to prove the usefulness of PACE in helping device design within an interested wavelength range.
We also test the accuracy outside the C-band, where \name shows good accuracy on neighboring wavelengths while holding a 10-15\% error at a further range. This is expected since wave propagation is sensitive to frequency. It can be mitigated by incorporating sampled wavelengths into training.

\noindent\textbf{Are \name a general enhancer module for Fourier-type operator?}
We further investigate whether our new \name operator is a general enhancer for other Fourier operators, rather than a dedicated module for our own model architecture.
We randomly insert four \name blocks into Facztoried FNO~\cite{ML_ICLR2023_Tran} and test the error on Metaline3x3 and Etched MMI 3x3 benchmarks, showing up to 28\% error re-
\begin{wrapfigure}{r}{0.4\textwidth}
\vspace{-5pt}
\includegraphics[width=\linewidth]{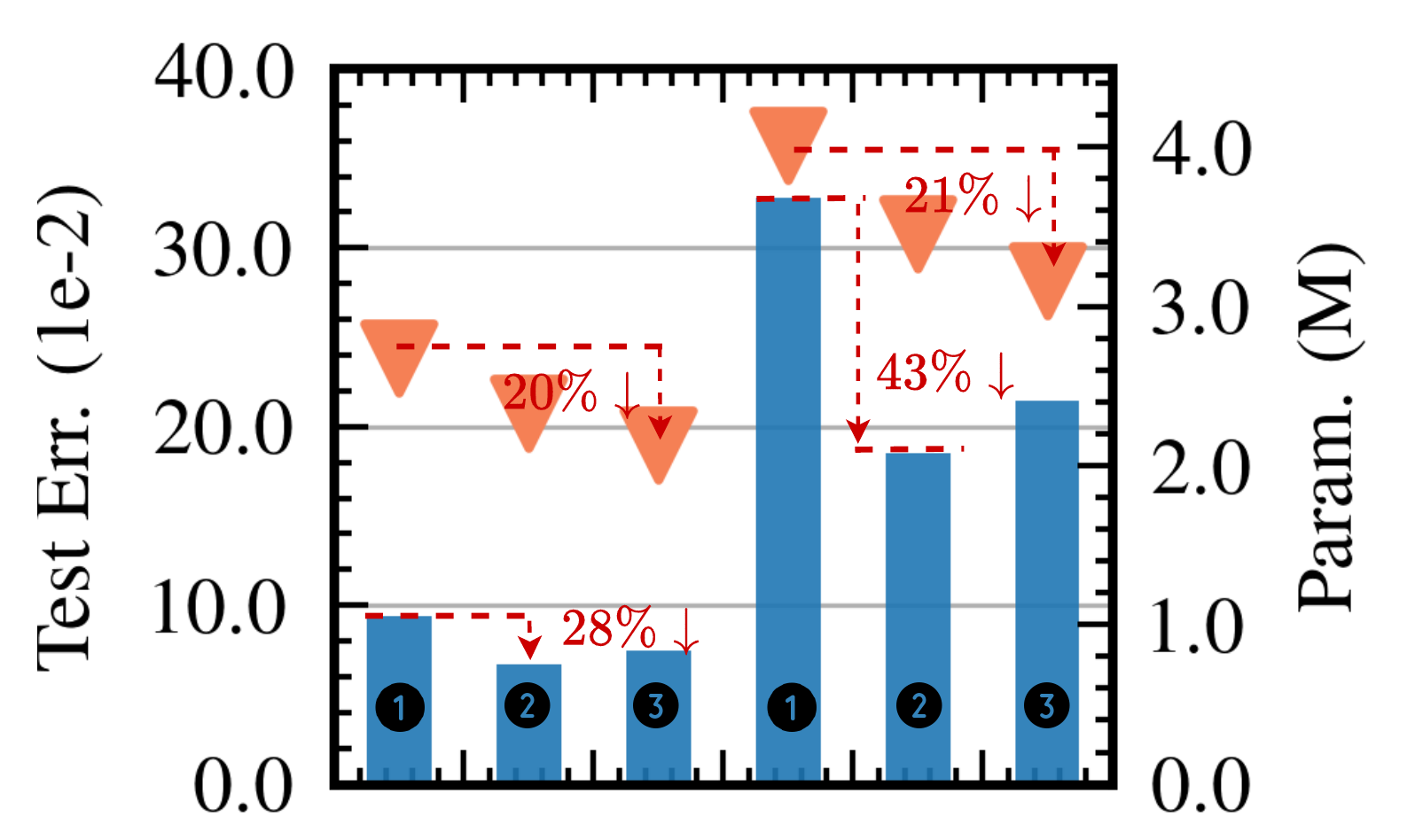} 
\caption{\small Insert 4 \name module (\ding{203}: g=2; \ding{204}: g=4) randomly in Factorized FNO(\ding{202}).}
\label{fig:insert}
\vspace{-30pt}
\end{wrapfigure}
duction as shown in Fig.~\ref{fig:insert} with much fewer parameters.

\noindent\textbf{Comparison with operator for multi-scale PDE}.
Noticing that our problem shares similar complexities in solving multi-scale PDEs with neural operator~\cite{liu2022ht, xu2023dilated},
we further compare our approach with the recent method~\cite{xu2023dilated} that alternates Fourier operator with dilated convolution layer to better capture local details.
On the etched MMI 3x3 dataset, we implement a 14-layer model with alternating \neurolight block and dilated convolution layer.
It yields a $1.73$ M parameter count similar to our \name but shows a $17.4$ N-MSE error, much worse than ours (10.59).

\begin{figure}
\vspace{-20pt}
\includegraphics[width=\linewidth]{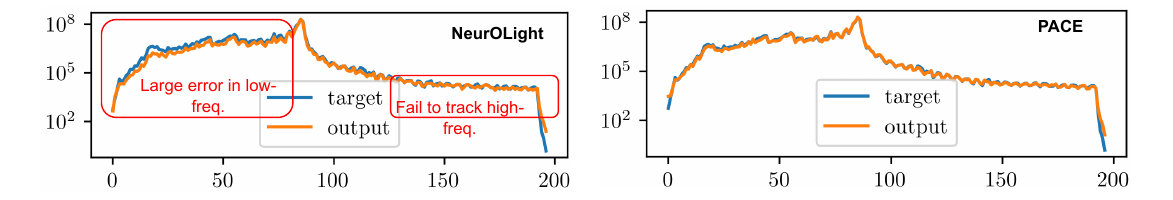}
    \caption{\small
     The radial energy spectrum of predicted fields from NeurOLight and PACE. NeurOLight fails to align precisely with the targeted field in both low-frequency and high-frequency parts.
    }
    \label{fig:spectrum}
    \vspace{-10pt}  
\end{figure}

\noindent\textbf{Spectrum of the predicted field}:
The predicted field spectrums of PACE and \neurolight are in Fig.~\ref{fig:spectrum}.
Although \neurolight uses the same frequency modes, it fails to align well with both the low-frequency and high-frequency regions.
\name excellently aligns with the baseline spectrum compared to NeurOLight, 

\section{Conclusion}
\label{sec:Conclusion}
In this work, we \textit{pace} the simulation fidelity on highly challenging \emph{complicated} photonic devices to an unprecedented level.
Our novel cross-axis factorized \name operator enables the neural PDE solver to capture complex relationships between local device structures and the resulting complex optical field across the entire simulation domain.
Furthermore, we introduce a cascaded two-stage learning paradigm to further enhance the prediction quality when one sole \name is not sufficient, demonstrating better quality enhancement than simply adding more layers.
Experiments demonstrate that \name achieves a remarkable 73\% reduction in error with 50\% fewer parameters compared to previous methods.  Our method also offers significant speedup (11.8x to 577x) over traditional numerical solvers.  Looking forward, we aim to integrate our model into the design optimization loop for photonic devices and circuits.  
Moreover, we want to emphasize that our proposed operator and learning strategy are not dedicated to photonic cases but generally applied to challenging PDE problems with similar problem characteristics, e.g., multi-scale PDE problems.  

\paragraph{Limitations and Broader Impact.} 
This work focuses on steady-state optical field solutions using the FDFD method. Exploring the effectiveness of operator learning for the Finite-Difference Time Domain (FDTD) can be an interesting direction.
Moreover, the FFT kernels on GPU are not fully optimized~\cite{fu2023flashfftconv}. Employing specialized, optimized FFT kernels can unlock even greater computational efficiency on GPUs, further accelerating the neural PDE solver.

\section{Acknowledgments and Disclosure of Funding}
We acknowledge NVIDIA for donating its A100 GPU workstations and the support from TILOS, NSF-funded National Artificial Intelligence Research Institute.
Additionally, this work was supported by the Air Force Office of Scientific Research (AFOSR) through the AFOSR project, contract FA9550-23-1-0452, and the Multidisciplinary University Research Initiative (MURI) program under contract No. FA9550-17-1-0071.
\newpage
\vspace{-.05in}
{
\bibliographystyle{plain}
\bibliography{./ref/Top_sim,./ref/Top,./ref/NN,./ref/NP}

\begin{thebibliography}{10}

\bibitem{ammari2012multi}
Habib Ammari, Yves Capdeboscq, and Hyeonbae Kang.
\newblock {\em Multi-scale and High-contrast PDE: from Modelling, to Mathematical Analysis, to Inversion}, volume 577.
\newblock American Mathematical Society, 2012.

\bibitem{ML_Arxiv2022_Ashiqur}
Md~Ashiqur~Rahman, Zachary~E Ross, and Kamyar Azizzadenesheli.
\newblock U-no: U-shaped neural operators.
\newblock {\em arXiv e-prints}, pages arXiv--2204, 2022.

\bibitem{cao2021choose}
Shuhao Cao.
\newblock Choose a transformer: Fourier or galerkin.
\newblock {\em Advances in neural information processing systems}, 34:24924--24940, 2021.

\bibitem{NP_Nature2021_Chen}
Mingkun Chen, Robert Lupoiu, Chenkai Mao, Der-Han Huang, Jiaqi Jiang, Philippe Lalanne, and Jonathan Fan.
\newblock Physics-augmented deep learning for high-speed electromagnetic simulation and optimization.
\newblock {\em Nature}, 2021.

\bibitem{NP_Nature2021_Feldmann}
Johannes Feldmann, Nathan Youngblood, Maxim Karpov, Helge Gehring, Xuan Li, Maik Stappers, Manuel~Le Gallo, Xin Fu, Anton Lukashchuk, Arslan Raja, Junqiu Liu, David Wright, Abu Sebastian, Tobias Kippenberg, Wolfram Pernice, and Harish Bhaskaran.
\newblock Parallel convolutional processing using an integrated photonic tensor core.
\newblock {\em Nature}, 2021.

\bibitem{feng2024integrated}
Chenghao Feng, Jiaqi Gu, Hanqing Zhu, Shupeng Ning, Rongxing Tang, May Hlaing, Jason Midkiff, Sourabh Jain, David~Z Pan, and Ray~T Chen.
\newblock Integrated multi-operand optical neurons for scalable and hardware-efficient deep learning.
\newblock {\em Nanophotonics}, 13(12):2193--2206, 2024.

\bibitem{NP_ACSPhotonics2022_Feng}
Chenghao Feng, Jiaqi Gu, Hanqing Zhu, Zhoufeng Ying, Zheng Zhao, David~Z Pan, and Ray~T Chen.
\newblock A compact butterfly-style silicon photonic-electronic neural chip for hardware-efficient deep learning.
\newblock {\em ACS Photonics}, 9(12):3906--3916, 2022.

\bibitem{fu2023flashfftconv}
Daniel~Y. Fu, Hermann Kumbong, Eric Nguyen, and Christopher R{\'e}.
\newblock Flash{FFTC}onv: Efficient convolutions for long sequences with tensor cores.
\newblock 2023.

\bibitem{gu2022squeezelight}
Jiaqi Gu, Chenghao Feng, Hanqing Zhu, Zheng Zhao, Zhoufeng Ying, Mingjie Liu, Ray~T Chen, and David~Z Pan.
\newblock Squeezelight: A multi-operand ring-based optical neural network with cross-layer scalability.
\newblock {\em IEEE Transactions on Computer-Aided Design of Integrated Circuits and Systems}, 42(3):807--819, 2022.

\bibitem{NP_Neurips2022_Gu}
Jiaqi Gu, Zhengqi Gao, Chenghao Feng, Hanqing Zhu, Ray Chen, Duane Boning, and David Pan.
\newblock Neurolight: A physics-agnostic neural operator enabling parametric photonic device simulation.
\newblock {\em Advances in Neural Information Processing Systems}, 35:14623--14636, 2022.

\bibitem{gu2024m3icro}
Jiaqi Gu, Hanqing Zhu, Chenghao Feng, Zixuan Jiang, Ray~T Chen, and David~Z Pan.
\newblock M3icro: Machine learning-enabled compact photonic tensor core based on programmable multi-operand multimode interference.
\newblock {\em APL Machine Learning}, 2(1), 2024.

\bibitem{NN_NeurIPS2021_Gupta}
Gaurav Gupta, Xiongye Xiao, and Paul Bogdan.
\newblock Multiwavelet-based operator learning for differential equations.
\newblock In {\em Proc.~NeurIPS}, 2021.

\bibitem{NN_arxiv2022_pdearena}
Jayesh~K Gupta and Johannes Brandstetter.
\newblock Towards multi-spatiotemporal-scale generalized pde modeling.
\newblock {\em arXiv preprint arXiv:2209.15616}, 2022.

\bibitem{NP_ACSPhotonics2018_Lim}
Tyler~W. Hughes, Momchil Minkov, Ian A.~D. Williamson, and Shanhui Fan.
\newblock Adjoint method and inverse design for nonlinear nanophotonic devices.
\newblock {\em ACS Photonics}, 2018.

\bibitem{NP_OE2023_Kim}
Junhyeong Kim, Berkay Neseli, Jae yong Kim, Jinhyeong Yoon, Hyeonho Yoon, Hyo hoon Park, and Hamza Kurt.
\newblock Inverse design of an on-chip optical response predictor enabled by a deep neural network.
\newblock {\em Opt. Express}, 2023.

\bibitem{TensorizedFNO}
Jean Kossaifi, Nikola Kovachki, Kamyar Azizzadenesheli, and Anima Anandkumar.
\newblock Multi-grid tensorized fourier neural operator for high-resolution pdes.
\newblock {\em arXiv preprint arXiv:2310.00120}, 2023.

\bibitem{li2023transformer}
Zijie Li, Kazem Meidani, and Amir~Barati Farimani.
\newblock Transformer for partial differential equations{\textquoteright} operator learning.
\newblock {\em Transactions on Machine Learning Research}, 2023.

\bibitem{ML_Neurips2024_Li}
Zijie Li, Dule Shu, and Amir Barati~Farimani.
\newblock Scalable transformer for pde surrogate modeling.
\newblock {\em Advances in Neural Information Processing Systems}, 36, 2024.

\bibitem{NN_ICLR2021_Li}
Zongyi Li, Nikola Kovachki, Kamyar Azizzadenesheli, Burigede Liu, Kaushik Bhattacharya, Andrew Stuart, and Anima Anandkumar.
\newblock Fourier neural operator for parametric partial differential equations.
\newblock In {\em International Conference on Learning Representations (ICLR)}, 2021.

\bibitem{NP_APL2022_Lim}
Joowon Lim and Demetri Psaltis.
\newblock Maxwellnet: Physics-driven deep neural network training based on maxwell’s equations.
\newblock {\em Appl. Phys. Lett.}, 2022.

\bibitem{liu2022ht}
Xinliang Liu, Bo~Xu, and Lei Zhang.
\newblock Ht-net: Hierarchical transformer based operator learning model for multiscale pdes.
\newblock 2022.

\bibitem{meinzer2014plasmonic}
Nina Meinzer, William~L Barnes, and Ian~R Hooper.
\newblock Plasmonic meta-atoms and metasurfaces.
\newblock {\em Nature photonics}, 8(12):889--898, 2014.

\bibitem{NP_JLT24_Ning}
Shupeng Ning, Hanqing Zhu, Chenghao Feng, Jiaqi Gu, Zhixing Jiang, Zhoufeng Ying, Jason Midkiff, Sourabh Jain, May~H Hlaing, David~Z Pan, et~al.
\newblock Photonic-electronic integrated circuits for high-performance computing and ai accelerators.
\newblock {\em Journal of Lightwave Technology}, 2024.

\bibitem{NP_ohlberger2018new}
Mario Ohlberger and Barbara Verfurth.
\newblock A new heterogeneous multiscale method for the helmholtz equation with high contrast.
\newblock {\em Multiscale Modeling \& Simulation}, 16(1):385--411, 2018.

\bibitem{NN_Neurips2024_Bogdan}
Bogdan Raonic, Roberto Molinaro, Tim De~Ryck, Tobias Rohner, Francesca Bartolucci, Rima Alaifari, Siddhartha Mishra, and Emmanuel de~B{\'e}zenac.
\newblock Convolutional neural operators for robust and accurate learning of pdes.
\newblock {\em Advances in Neural Information Processing Systems}, 36, 2024.

\bibitem{NP_NaturePhotonics2021_Shastri}
Bhavin~J. Shastri, Alexander~N. Tait, T.~Ferreira de~Lima, Wolfram H.~P. Pernice, Harish Bhaskaran, C.~D. Wright, and Paul~R. Prucnal.
\newblock {Photonics for Artificial Intelligence and Neuromorphic Computing}.
\newblock {\em Nature Photonics}, 2021.

\bibitem{shi2022silicon}
Yaocheng Shi, Yong Zhang, Yating Wan, Yu~Yu, Yuguang Zhang, Xiao Hu, Xi~Xiao, Hongnan Xu, Long Zhang, and Bingcheng Pan.
\newblock Silicon photonics for high-capacity data communications.
\newblock {\em Photonics Research}, 10(9):A106--A134, 2022.

\bibitem{NN_ICLR2021_Kim}
Kim Stachenfeld, Drummond~Buschman Fielding, Dmitrii Kochkov, Miles Cranmer, Tobias Pfaff, Jonathan Godwin, Can Cui, Shirley Ho, Peter Battaglia, and Alvaro Sanchez-Gonzalez.
\newblock Learned simulators for turbulence.
\newblock In {\em International conference on learning representations}, 2021.

\bibitem{NP_SciRep2019_Tahersima}
Mohammad~H. Tahersima, Keisuke Kojima, Toshiaki Koike-Akino, Devesh Jha, BingnanWang, and Chungwei Lin.
\newblock Deep neural network inverse design of integrated photonic power splitters.
\newblock {\em Sci. Rep.}, 2019.

\bibitem{ML_ICLR2023_Tran}
Alasdair Tran, Alexander Mathews, Lexing Xie, and Cheng~Soon Ong.
\newblock Factorized fourier neural operators.
\newblock {\em arXiv preprint arXiv:2111.13802}, 2021.

\bibitem{NN_NeurIPSW2021_Tran}
Alasdair Tran, Alexander Mathews, Lexing Xie, and Cheng~Soon Ong.
\newblock Factorized fourier neural operators.
\newblock In {\em NeurIPS Workshop}, 2021.

\bibitem{NN_ICLR2023_Tran}
Alasdair Tran, Alexander Mathews, Lexing Xie, and Cheng~Soon Ong.
\newblock Factorized fourier neural operators.
\newblock In {\em The Eleventh International Conference on Learning Representations}, 2023.

\bibitem{NP_SciRep2019_Trivedi}
Rahul Trivedi, Logan Su, Jesse Lu, Martin~F. Schubert, and JelenaVuckovic.
\newblock Data-driven acceleration of photonic simulations.
\newblock {\em Sci. Rep.}, 2019.

\bibitem{verfurth2019heterogeneous}
Barbara Verf{\"u}rth.
\newblock Heterogeneous multiscale method for the maxwell equations with high contrast.
\newblock {\em ESAIM: Mathematical Modelling and Numerical Analysis}, 53(1):35--61, 2019.

\bibitem{wang2022integrated}
Zi~Wang, Lorry Chang, Feifan Wang, Tiantian Li, and Tingyi Gu.
\newblock Integrated photonic metasystem for image classifications at telecommunication wavelength.
\newblock {\em Nature communications}, 13(1):2131, 2022.

\bibitem{ML_ICML2023_Wu}
Haixu Wu, Tengge Hu, Huakun Luo, Jianmin Wang, and Mingsheng Long.
\newblock Solving high-dimensional pdes with latent spectral models.
\newblock In {\em Proceedings of the 40th International Conference on Machine Learning}, ICML'23. JMLR.org, 2023.

\bibitem{xiong2020layer}
Ruibin Xiong, Yunchang Yang, Di~He, Kai Zheng, Shuxin Zheng, Chen Xing, Huishuai Zhang, Yanyan Lan, Liwei Wang, and Tieyan Liu.
\newblock On layer normalization in the transformer architecture.
\newblock In {\em International Conference on Machine Learning}, pages 10524--10533. PMLR, 2020.

\bibitem{xu2023dilated}
Bo~Xu and Lei Zhang.
\newblock Dilated convolution neural operator for multiscale partial differential equations.
\newblock 2023.

\bibitem{NP_Nature2021_Xu}
Xingyuan Xu, Mengxi Tan, Bill Corcoran, Jiayang Wu, Andreas Boes, Thach~G. Nguyen, Sai~T. Chu, Brent~E. Little, Damien~G. Hicks, Roberto Morandotti, Arnan Mitchell, and David~J. Moss.
\newblock {11 TOPS photonic convolutional accelerator for optical neural networks}.
\newblock {\em Nature}, 2021.

\bibitem{zarei2020integrated}
Sanaz Zarei, Mahmood-reza Marzban, and Amin Khavasi.
\newblock Integrated photonic neural network based on silicon metalines.
\newblock {\em Optics Express}, 28(24):36668--36684, 2020.

\bibitem{NP_NatureComm2022_Zhu}
H.~H. Zhu, J.~Zou, H.~Zhang, Y.~Z. Shi, S.~B. Luo, et~al.
\newblock Space-efficient optical computing with an integrated chip diffractive neural network.
\newblock {\em Nature Communications}, 2022.

\bibitem{NP_HPCA24_Zhu}
Hanqing Zhu, Jiaqi Gu, Hanrui Wang, Zixuan Jiang, Zhekai Zhang, Rongxing Tang, Chenghao Feng, Song Han, Ray~T Chen, and David~Z Pan.
\newblock Lightening-transformer: A dynamically-operated optically-interconnected photonic transformer accelerator.
\newblock In {\em 2024 IEEE International Symposium on High-Performance Computer Architecture (HPCA)}, pages 686--703. IEEE, 2024.

\bibitem{zhu2022fuse}
Hanqing Zhu, Keren Zhu, Jiaqi Gu, Harrison Jin, Ray~T Chen, Jean~Anne Incorvia, and David~Z Pan.
\newblock Fuse and mix: Macam-enabled analog activation for energy-efficient neural acceleration.
\newblock In {\em Proceedings of the 41st IEEE/ACM International Conference on Computer-Aided Design}, pages 1--9, 2022.

\end{thebibliography}
}

\newpage
\appendix

\section{Appendix}

\subsection{Dataset Generation}
\label{subsec:dataset}
We generate our customized etched MMI and Metaline dataset using the open-source FDFD simulator angler~\cite{NP_ACSPhotonics2018_Lim}.
For each type of device, we random sample 5.12 K device configuration following the Tab.~\ref{tab:AppendixDataset}, and generate single-source data by sweeping the input light over the input ports.
We randomly sample the device's physical dimension, input/output waveguide width and input light source frequencies.
For etched MMIs, we randomly sample etched cavity sizes, ratios (which determine the number of cavities in the MMIs), and permittivities in the controlling region.
For Metaline, we randomly sample the metaatom physical dimension with a fixed total number of 20.

We discretize the domain of etched MMI by $80\times 384$, and the domain of metaline by $128\times 144$.

\begin{table}[htp!]
\centering
\caption{Summary of etched MMI device design variable’s sampling range, distribution, and unit.}
\label{tab:AppendixDataset}
\resizebox{0.8\columnwidth}{!}{%
\begin{tabular}{l|ccc|c}
\hline
\multirow{2}{*}{Variables} & \multicolumn{3}{c|}{Value/Distribution}     & \multirow{2}{*}{Unit} \\ \cmidrule{2-4}
\multicolumn{1}{c|}{}                           &  Metaline $3\times3$           &  Etched MMI $3\times3$         &  Etched MMI $5\times5$ & \multicolumn{1}{c}{}                      \\ \midrule 
Length                                          & $\mathcal{U}($8, 10)          & $\mathcal{U}($20, 30)            & $\mathcal{U}($25, 35)                        & $\mu m$                                        \\
Width                                           & $\mathcal{U}($10, 12)          & $\mathcal{U}($5.5, 7)            & $\mathcal{U}($7.5, 9)                       & $\mu m$                                        \\
Port Length                                     & 1.5                               & 1.5                           & 1.5                               & $\mu m$                                        \\
Port Width                                      & $\mathcal{U}($0.5, 0.8)                   & $\mathcal{U}($0.8, 1.1)          & $\mathcal{U}($0.8, 1.1)                     & $\mu m$                                        \\
Taper Length                                    & 3                 & 4.5                    & 4.5                               & $\mu m$                                        \\
Taper Width                                     & 1.3                  & 1.3                    & 1.3                               & $\mu m$                                        \\
Border Width                 & 0.25                    & 0.25                 & 0.25                            & $\mu m$                                        \\
PML Width                                       & 1.5                  & 1.5                  & 1.5                             & $\mu m$                                        \\
Wavelengths $\lambda$                           & $\mathcal{U}($1.53, 1.565)           & $\mathcal{U}($1.53, 1.565)       & $\mathcal{U}($1.53, 1.565)                  & $\mu m$                                        \\
Cavity Ratio                                    & -             & $\mathcal{U}($0.05, 0.1)                    & $\mathcal{U}($0.05, 0.1)                    & -                                         \\
Note                                    & 2-layer random meta-atoms             & random slots                    & random slots                   & -                                         \\
Relative Permittivity $\eps_r$                  & \{2.07, 12.11\}                  & \{2.07, 12.11\}        & \{2.07, 12.11\}                          & -                                         \\ \bottomrule
\end{tabular}%
}
\end{table}

\subsection{Training Settings}

We implement all models and training logic in PyTorch 2.3. 
We use A100 and A6000 to train our models and report the latency running on a single A100 GPU with \texttt{torch.compile}.
For Benchmarking FDFD simulator performance, we use the Intel 12th Gen Intel(R) Core(TM) i7-12700 with 20 CPU cores.
We split all single-source examples into 72\% training data, 8\% validation data, and 20\% test data.

For training, we set the number of epochs to 100 with an initial learning rate of 0.002, cosine learning rate decay, and a mini-batch size of 4.
We use adamW as the optimizer with the weight decay 1e-5 to avoid over-fitting.
Moreover, we apply stochastic network depth with a linear scaling strategy.

\subsection{Model Designs}
\label{subsec:model_details}

To ensure a comprehensive evaluation, we compare our proposed model against recently published and available SOTA baselines, encompassing various architectural paradigms such as the Fourier-operator models, attention-based models, and latent space methods.
To maintain fairness in comparison, we constrain the parameter count of all models to be under 4 M in most cases and use open-sourced implementations.

Here, we report the model details for baselines for the etched MMI dataset.

\noindent\textbf{UNet}.
We construct a 4-level convolutional UNet with a base channel number of 36, following the open-sourced implementation\footnote{https://github.com/JeremieMelo/NeurOLight}.
The total parameter count is 3.88 M.

\noindent\textbf{Dil-ResNet}~\cite{NN_ICLR2021_Kim}.
We use the implementation in open-sourced pdearena~\cite{NN_arxiv2022_pdearena}\footnote{https://pdearena.github.io/pdearena/}, with a channel number of 128 and enabled normalization. The total parameter count for Dil-ResNet is 4.17 million.

\noindent\textbf{FNO-2d}~\cite{NN_ICLR2021_Li}.
We use 6 2-D FNO layers, and the Fourier modes are set to (\#Mode =32, \#Mode =10) for the etched MMI dataset and (\#Mode =16, \#Mode =16) for the Metaline dataset, resulting in the total parameter count as 3.99 M or 3.21 M.
We use the implementation\footnote{https://github.com/JeremieMelo/NeurOLight}.

\noindent\textbf{Tensorized FNO-2d}~\cite{TensorizedFNO}.
One obvious advantage is that our model features low parameters. 
Hence, we will compare it with tensorized FNO, which compresses the model weights with the tensor decomposition method.
We adopt the implementation in\footnote{https://github.com/neuraloperator/neuraloperator} and use the model designed for the darcy flow problem.
We use 5 2-D FNO layers, and the Fourier modes are set to (\#Mode =40, \#Mode =20) for the etched MMI dataset and (\#Mode =24, \#Mode =24) for the Metaline dataset.
Tucker decomposition is used with a rank of 0.42.
The total parameter count is 2.25 M and 1.58 M, respectively, for the two types of datasets.

\noindent\textbf{F-FNO-2d}.
For factorized Fourier neural operator (F-FNO), we use 13 F-FNO layers with a channel number of 52. The Fourier modes are set to (\#Mode =70, \#Mode =40) for etched MMI dataset and (\#Mode =36, \#Mode =36) for Metaline dataset, leading to the total parameter count as 4.02 M and 2.68M.
The FNO-2d implementation is referred to\footnote{https://github.com/alasdairtran/fourierflow}.
We use the same projection head as ours.

\noindent\textbf{U-NO-2d}~\cite{ML_Arxiv2022_Ashiqur}.
For U-shaped neural operators, we follow the implementation\footnote{https://github.com/ashiq24/UNO/blob/main/navier\_stokes\_uno2d.py}.
We use their 11-layer UNO with a base channel 24.
The total parameter count is 4.38 M.

\noindent\textbf{Attention-based operator}~\cite{ML_Neurips2024_Li}.
For the attention-based neural operator, we choose the most recent Transformer-type model~\cite{ML_Neurips2024_Li} and use its official implementation\footnote{https://github.com/BaratiLab/FactFormer/tree/main}.
We use 3 layer-attention with 12 heads. The total dimension is 128.
The total parameter count is 3.75 M.

\noindent\textbf{Latent Spectral Method}~\cite{ML_ICML2023_Wu}.
For the latent spectral method, we use the original implementation in\footnote{https://github.com/thuml/Latent-Spectral-Models}.
The number of bases is set to 12, and the channel number is 32.
The patch size is set to 4$\times$4.
The total parameter count is 4.8 M.

\noindent\textbf{\ffno}~\cite{NP_Neurips2022_Gu}.
We use the same implementation\footnote{https://github.com/JeremieMelo/NeurOLight} in the original paper with 16 layers. 
For the etched MMI dataset,
the Fourier modes are set to (\#Mode =70, \#Mode =40).
For the Metaline dataset, Fourier modes are set to (\#Mode =36, \#Mode =36).
The total number of parameters is 2.11M and 1.49 M for the two cases.

\noindent\textbf{\name}.
For our proposed \name, we use 12 layers, with the first two being the same factorized layers in~\cite{NP_Neurips2022_Gu}, since we found it is important first to generate some meaningful wave patterns and then do global information swapping. 
The Fourier modes are set to (\#Mode =70, \#Mode =40).
We use the same convolution stem in ~\cite{NP_Neurips2022_Gu} to extract information before going through the feature propagator and the same projection head. 
The total number of parameters is 1.73M.
For the second stage \nameb model, we use 8 layers with all being \name operators, where Fourier modes are set to (\#Mode =100, \#Mode =40) 
For the Metaline dataset, we solely use a 12-layer \name with Fouier modes being (\#Mode =36, \#Mode =36).

\subsection{Ablation study on group number choices}
\label{subsec:group}

We run an 8-layer \name model on the Metaline dataset by setting the group size to 1, 2, 4, and show the train and test error in Tab.~\ref{tab:ModelGroup}
We use \#group=4 in our paper, which balances between parameter efficiency and test error.

\begin{table}
\centering
\caption{Ablation on \# group on an 8-layer \name model on the Metaline dataset.}
\label{tab:ModelGroup}
\resizebox{0.5\linewidth}{!}{%
\begin{tabular}{@{}cccc@{}}
\toprule
\# Group & \begin{tabular}[c]{@{}c@{}}\#Params \\ (M)$\downarrow$\end{tabular}  & \begin{tabular}[c]{@{}c@{}}\#Train Err \\ ($10^{-2}$)$\downarrow$\end{tabular}  & \begin{tabular}[c]{@{}c@{}}\#Test Err \\ ($10^{-2}$)$\downarrow$\end{tabular}  \\ \midrule
1 & 2.15 & 4.89 & 4.55 \\
2 & 1.27 & 5.33 & 4.80 \\
4 & 0.825 & 5.65 & 4.82 \\  
\bottomrule
\end{tabular}
}
\end{table}

\subsection{Ablation Study of Double Skip and Pre-Normalization}

We further investigate whether the observed improvements in accuracy are attributed to the incorporation of double skip connections and pre-normalization, which were incorporated into our model to stabilize it in deeper layers with better generalization. 
We add these two techniques to \neurolight and compare them with ours \name in Tab..~\ref{tab:doubleskip}.
The double skip and pre-normalization can make the model generalize well for test data, while the training error is slightly improved as normalization can be seen as some linear affine.
However, it still shows much worse accuracy than our model, especially given the context our model is shallower with fewer parameters.

\begin{table}
\centering
\caption{Ablation on the comparison between \name and \neurolight with the adopted double skip and pre-normalization.}
\label{tab:doubleskip}
\resizebox{\linewidth}{!}{%
\begin{tabular}{@{}ccccc@{}}
\toprule
Model & Double skip \& Pre-norm. & \begin{tabular}[c]{@{}c@{}}\#Params \\ $\downarrow$\end{tabular}  & \begin{tabular}[c]{@{}c@{}}\#Train Err \\ ($10^{-2}$)$\downarrow$\end{tabular}  & \begin{tabular}[c]{@{}c@{}}\#Test Err \\ ($10^{-2}$)$\downarrow$\end{tabular}  \\ \midrule
\neurolight-16 layer & \ding{53} & 2108258 & 15.58 & 17.21 \\
\neurolight-16 layer & \ding{52} & 2110306 & 15.26 & 15.87 \\
\name-12 layer & \ding{53} & 1709026 & 10.32 & 11.06 \\
\name-12 layer & \ding{52} & 1710562 & 9.60 & 10.60 \\
\bottomrule
\end{tabular}
}
\end{table}

\subsection{L2 distance is a more informative metric compared to L1 distance for distance evaluation}
\label{subsec:mse}

\begin{corollary}\label{cor:dist}
Consider two complex numbers in polar form, $z_1 = r_1 \angle \phi_1$ and $z_2 = r_2 \angle \phi_2$. Their mean square error is rotation invariant, as shown by:
\begin{equation}
\label{eq:l2distance}
\begin{aligned}
| z_{1}, z_{2}|_{2}^2 &= | r_1 \cos \phi_1 -r_2 \cos \phi_2|^{2} + | r_1 \sin \phi_1 -r_2 \sin \phi_2|^{2} \
&= r_1^2 + r_2^2 - 2 r_1 r_2 \cos (\phi_1 - \phi_2).
\end{aligned}
\end{equation}
This distance metric depends solely on the difference in signal norm and angle between $z_1$ and $z_2$. However, their mean absolute error is the rotation variant:
\begin{equation}
\label{eq:l1distance}
\begin{aligned}
| z{1}, z_{2}|_{1} &= | r_1 \cos \phi_1 -r_2 \cos \phi_2| + | r_1 \sin \phi_1 -r_2 \sin \phi_2|.
\end{aligned}
\end{equation}
\end{corollary}

\begin{figure}
\centering
\includegraphics[width=0.3\linewidth]{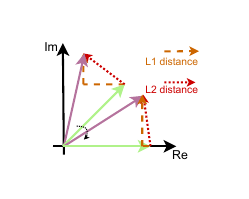} 
\caption{L1 and L2 distance in the complex plane.}
\label{fig:ComplexErr}
\end{figure}
In the complex plane, optimizing MAE equates to minimizing the summed L1 distances of the real and imaginary components.
However, the L1 distance is rotation-variant.
A simple rotation of the two complex numbers on the plane results in changing L1 distance, as shown in Fig.~\ref{fig:ComplexErr}, while the true distance does not alter.
Therefore, it is not an appropriate metric as it cannot accurately measure proximity in the complex plane. 
In this way, we use L2 distance in the loss (mean squared loss) that is rotation invariant, as proved in corollary~\ref{cor:dist}, which exactly captures the distance in the complex plane.

\subsection{Train two-stage model sequentially}
\label{subsec:seq}

We also show the error by training our proposed two-stage model sequentially in Tab.~\ref{tab:2stageSeqResults}, which shows a similar error to our joint training approach when equipping with our proposed cross-stage feature distillation.

Training sequentially is more costly than joint training, as second-stage training requires first inferencing with the first stage to get the predicted results.

\subsection{Visualization of energy spectrum}

\begin{figure}
\centering
\includegraphics[width=0.5\linewidth]{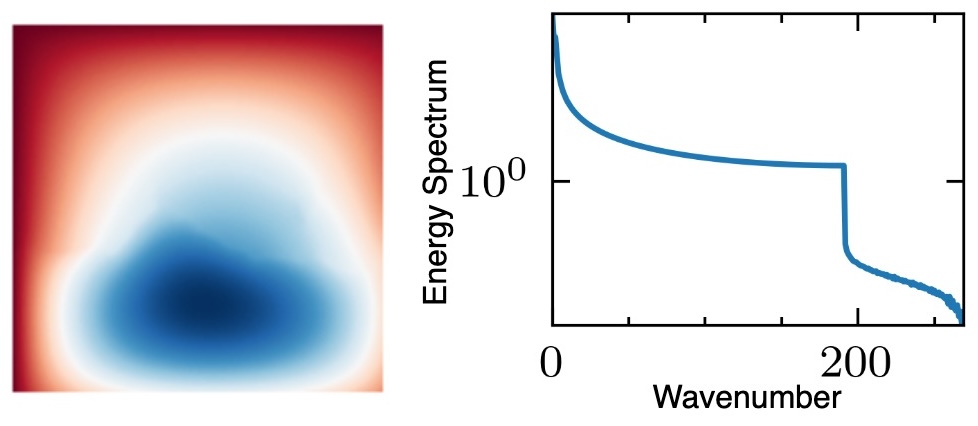} 
\caption{Radial energy spectrum for one solution of Darcy flow problem.}
\label{fig:darcy}
\end{figure}

\begin{table}
    \centering
    \captionof{table}{\small Results of train two-stage model sequentially.}
    \label{tab:2stageSeqResults}
    \resizebox{\textwidth}{!}{%
    \begin{tabular}{ccc|ccc}
    \toprule
    Benchmarks                    & Model      & Cross-stage dist.        & \multicolumn{1}{l}{\#Params (M) $\downarrow$} & \multicolumn{1}{c}{Train Err ($10^{-2}$) $\downarrow$}                         & \multicolumn{1}{c}{Test Err ($10^{-2}$) $\downarrow$}                          \\ \midrule
                                  & \cellcolor[HTML]{C0C0C0}\name-12 layer             &    -                          & \cellcolor[HTML]{C0C0C0}1.73                                 & \cellcolor[HTML]{C0C0C0}9.51                                 & \cellcolor[HTML]{C0C0C0} 10.59                                  \\ \cmidrule{2-6}
                                  \multirow{-1}{*}{Etched MMI 3x3} &  \namea $\rightarrow$ \nameb          &  \ding{52}                             &  3.151                                  &  3.91                                 &  5.06                                  \\
     \midrule
    & \cellcolor[HTML]{C0C0C0}\name-12 layer             &    -                          & \cellcolor[HTML]{C0C0C0}1.73                                 & \cellcolor[HTML]{C0C0C0} 11.66                                & \cellcolor[HTML]{C0C0C0} 11.91                                 \\ \cmidrule{2-6}                          
                                 \multirow{-1}{*}{Etched MMI 5x5} &  \namea $\rightarrow$ \nameb          &  \ding{52}                             &  3.151                                  &  5.51                                &  6.15                                  \\  
    \bottomrule
    \end{tabular}%
    }
\end{table}

We generate the prediction field's radial energy spectrum by first transferring the image from the spatial domain to the spectral domain and then shifting the transferred image to the center.
Then, the wavenumber is computed as the distance with respect to the center. 

We sum the squared magnitude of the Fourier coefficients that fall into the specific number, which is implemented following open-sourced code~\footnote{https://github.com/autonomousvision/projected\-gan/blob/main/torch\_utils/utils\_spectrum.py}.

We also visualize one example of darcy flow problem, as shown in Fig.~\ref{fig:darcy}.
It shows highly distant characteristics compared to our optical field, with most information concentrating on low-frequency parts.

\subsection{Visualization of feature map before/after non-linear activation in our explicitly designed high-frequency projection path}

\begin{figure}
    \centering
    \includegraphics[width=\textwidth]{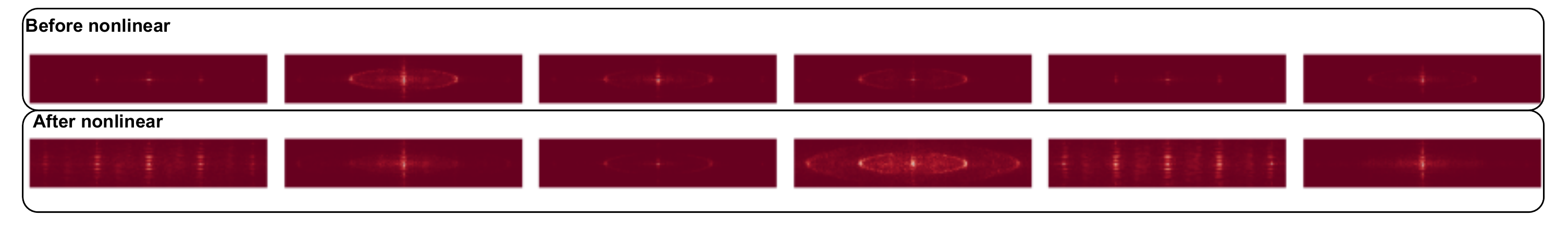}
    \caption{~\small
    Frequency-domain visualization of feature map before and after non-linear activation in the last PACE block(The center represents low frequency). The pattern is shifted to the center to understand the frequency content better.
    }
    \label{fig:high_freq}
\end{figure}

We visualize the first 6 channels of feature maps before and after the nonlinear activation in the last PACE layer by showing them in the frequency domain. As shown in Fig. ~\ref{fig:high_freq}, the nonlinear activation can ignite high-frequency features, which confirms our claim and validates our design choice of injecting an extra high-frequency projection path.

\subsection{Visualization of prediction}
\label{subsec: visual}
\begin{figure}
    \centering
    \includegraphics[width=\textwidth]{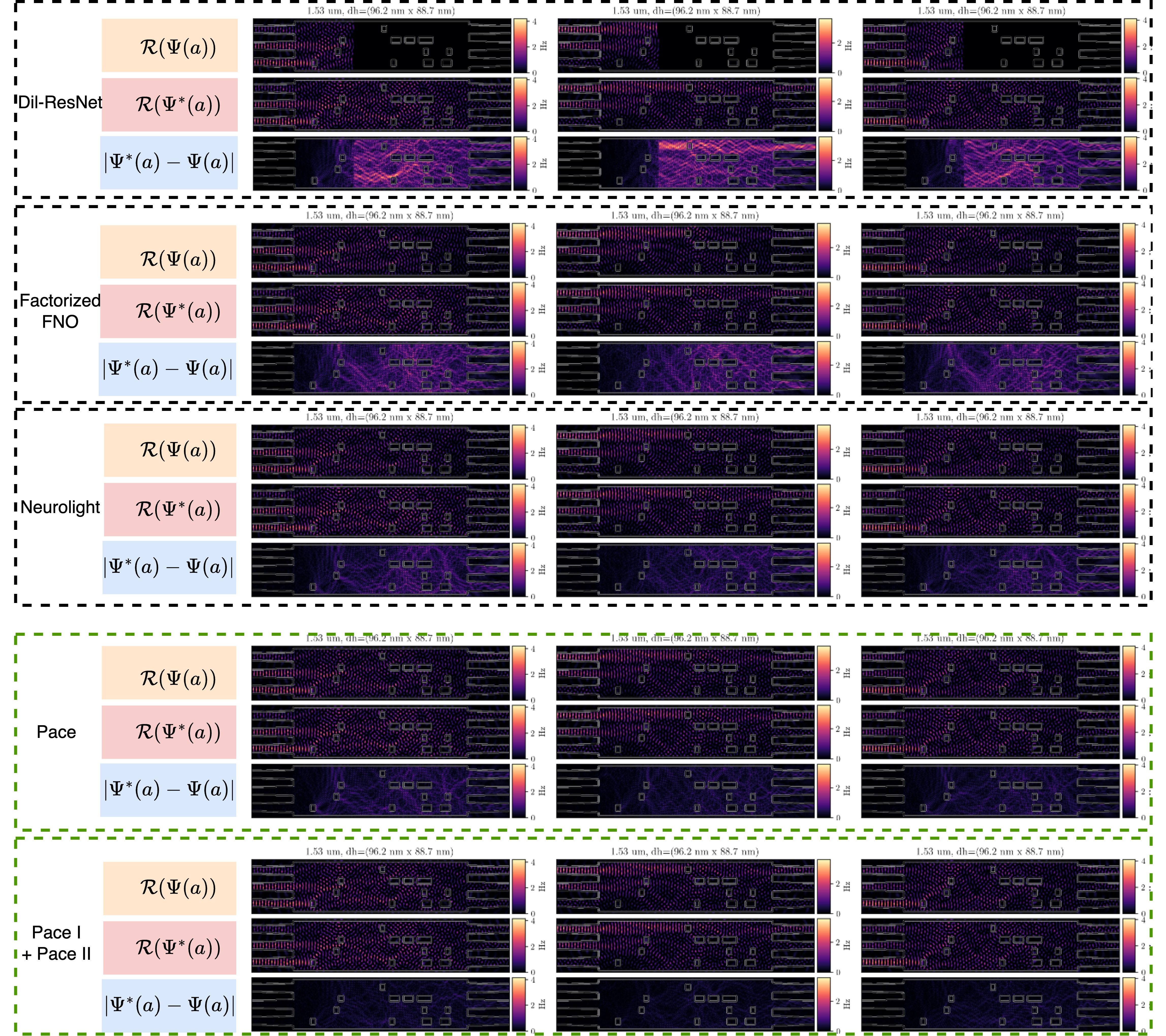}
    \caption{Visualization of test cases on etched MMI 3x3 devices with random input sources. }
    \label{fig:vis_mmi}
\end{figure}

\begin{figure}
    \centering
    \includegraphics[width=\textwidth]{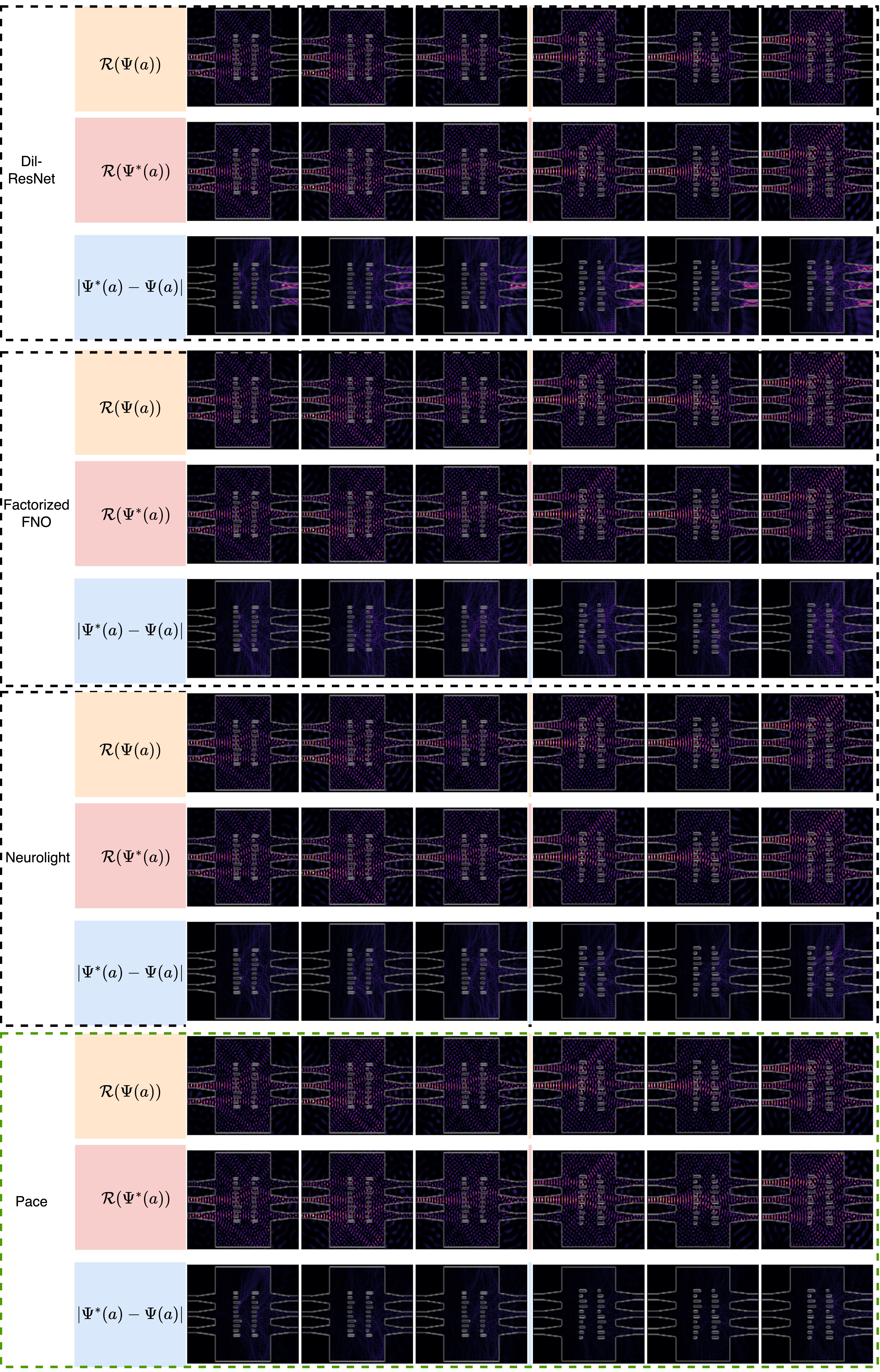}
    \caption{Visualization of several test cases on Metaline devices with random input sources.}
    \label{fig:vis_metaline}
\end{figure}

We provide visualization figures on etched MMI 3x3 devices in Fig.~\ref{fig:vis_mmi} and metaline devices in Fig.~\ref{fig:vis_metaline}.
We provide the predicted fields $\Psi(a)$, the groud-truth field $\Psi(a)^{*}$ and the residual error $\Psi(a)^* - \Psi(a)$ of Dil-ResNet, Facztoried FNO, \neurolight and our \name.
For etched MMI test cases, we show both the single 12-layer \name model and the joint 20-layer model \namea + \nameb.
Our \name shows much better prediction results with a near-black error map compared to other baseline methods.

\newpage

\clearpage
\section*{NeurIPS Paper Checklist}


\begin{enumerate}

\item {\bf Claims}
    \item[] Question: Do the main claims made in the abstract and introduction accurately reflect the paper's contributions and scope?
    \item[] Answer: \answerYes{} 

\item {\bf Limitations}
    \item[] Question: Does the paper discuss the limitations of the work performed by the authors?
    \item[] Answer: \answerYes{} 

\item {\bf Theory Assumptions and Proofs}
    \item[] Question: For each theoretical result, does the paper provide the full set of assumptions and a complete (and correct) proof?
    \item[] Answer:\answerYes{} 

    \item {\bf Experimental Result Reproducibility}
    \item[] Question: Does the paper fully disclose all the information needed to reproduce the main experimental results of the paper to the extent that it affects the main claims and/or conclusions of the paper (regardless of whether the code and data are provided or not)?
    \item[] Answer:\answerYes{} 
    
\item {\bf Open access to data and code}
    \item[] Question: Does the paper provide open access to the data and code, with sufficient instructions to faithfully reproduce the main experimental results, as described in supplemental material?
    \item[] Answer:\answerYes{} 
    \item[] Justification: We open-source the dataset and code.
    \item[] Guidelines:

\item {\bf Experimental Setting/Details}
    \item[] Question: Does the paper specify all the training and test details (e.g., data splits, hyperparameters, how they were chosen, type of optimizer, etc.) necessary to understand the results?
    \item[] Answer:\answerYes{} 
    
\item {\bf Experiment Statistical Significance}
    \item[] Question: Does the paper report error bars suitably and correctly defined or other appropriate information about the statistical significance of the experiments?
    \item[] Answer:\answerYes{} 
    \item[] Justification: We report the error bar when we test the performance on generalization experiments.

\item {\bf Experiments Compute Resources}
    \item[] Question: For each experiment, does the paper provide sufficient information on the computer resources (type of compute workers, memory, time of execution) needed to reproduce the experiments?
    \item[] Answer:\answerYes{} 
    
\item {\bf Code Of Ethics}
    \item[] Question: Does the research conducted in the paper conform, in every respect, with the NeurIPS Code of Ethics \url{https://neurips.cc/public/EthicsGuidelines}?
    \item[] Answer:\answerYes{} 

\item {\bf Broader Impacts}
    \item[] Question: Does the paper discuss both potential positive societal impacts and negative societal impacts of the work performed?
    \item[] Answer:\answerYes{} 

\item {\bf Safeguards}
    \item[] Question: Does the paper describe safeguards that have been put in place for responsible release of data or models that have a high risk for misuse (e.g., pretrained language models, image generators, or scraped datasets)?
    \item[] Answer:\answerNA{} 

\item {\bf Licenses for existing assets}
    \item[] Question: Are the creators or original owners of assets (e.g., code, data, models), used in the paper, properly credited and are the license and terms of use explicitly mentioned and properly respected?
    \item[] Answer:\answerNA{} 

\item {\bf New Assets}
    \item[] Question: Are new assets introduced in the paper well documented and is the documentation provided alongside the assets?
    \item[] Answer:\answerYes{} 
    \item[] Guidelines:

\item {\bf Crowdsourcing and Research with Human Subjects}
    \item[] Question: For crowdsourcing experiments and research with human subjects, does the paper include the full text of instructions given to participants and screenshots, if applicable, as well as details about compensation (if any)? 
    \item[] Answer:\answerNA{} 

\item {\bf Institutional Review Board (IRB) Approvals or Equivalent for Research with Human Subjects}
    \item[] Question: Does the paper describe potential risks incurred by study participants, whether such risks were disclosed to the subjects, and whether Institutional Review Board (IRB) approvals (or an equivalent approval/review based on the requirements of your country or institution) were obtained?
    \item[] Answer:\answerNA{} 

\end{enumerate}

\end{document}